# From Days to Minutes: An Autonomous AI Agent Achieves Reliable Clinical Triage in Remote Patient Monitoring

**Authors:** Seunghwan Kim, PhD[1], Tiffany H. Kung, MD[1,2], Heena Verma[1], Dilan Edirisinghe[1], Kaveh Sedehi, MBA[1], Johanna Alvarez, MSN, FNP-BC[1], Diane Shilling, DNP, FNP-C[1], Audra Lisa Doyle, MSN, FNP-BC[1], Ajit Chary, MD[1], William Borden, MD[1,3], Ming Jack Po, MD, PhD[1]

[1]AnsibleHealth Inc., San Francisco, CA, USA

[2]Department of Anesthesiology, Perioperative and Pain Medicine, Stanford School of Medicine, Stanford, CA

[3]Department of Medicine, Division of Cardiology, George Washington University, Washington, DC

**Correspondence:** nigel.kim@ansiblehealth.com

# Abstract

**Background.** Remote patient monitoring (RPM) generates vast quantities of vital sign data, yet landmark heart failure trials—including Tele-HF, BEAT-HF, and TIM-HF1—failed to improve outcomes because data floods overwhelmed clinical staff without intelligent filtering or contextual interpretation. TIM-HF2 proved that structured, responsive monitoring with 24/7 physician staffing reduces mortality by 30%, but this model is prohibitively expensive and unscalable.

**Methods.** We developed Sentinel, an autonomous AI agent using Model Context Protocol (MCP) to perform contextual clinical triage of RPM vital signs. The system equips a large language model with 21 structured clinical tools for comprehensive patient context retrieval, enabling multi-step clinical reasoning for each reading. We evaluated Sentinel through three studies: (1) agent self-consistency via inter-rater reliability (100 readings × 5 runs, Fleiss' κ); (2) comparison against rule-based threshold baselines on 500 readings from 340 patients; and (3) validation against 6 human clinicians (3 physicians, 3 nurse practitioners) using a connected matrix design (3 reviewers per sample). A leave-one-out (LOO) analysis compared the agent against each individual clinician, and all agent overtriage cases underwent independent clinical adjudication.

**Results.** The agent achieved 95.8% sensitivity for emergency classifications (23/24) and 88.5% sensitivity for all actionable alerts (emergency + urgent; 92/104), with 85.7% specificity (311/363) against the human majority-vote reference standard (N=467, ties excluded). Four-level exact accuracy was 69.4% (324/467) with a quadratic-weighted kappa of 0.778 (95% CI: 0.728–0.819), and 95.9% of classifications fell within one severity level. In leave-one-out analysis, the agent outperformed every individual clinician in both emergency sensitivity (97.5% vs. clinician aggregate 60.0%) and actionable sensitivity (90.9% vs. 69.5%). Agent-human disagreements skewed toward overtriage (22.5%) rather than undertriage (8.1%); independent physician adjudication of the 17 most severe overtriage cases (agent–majority gap ≥2 levels) found that both reviewers independently validated the agent's escalation in 88–94% of cases, with true overtriage confirmed in only 6–12%; subsequent consensus resolution between the two reviewers validated all agent's escalations (100% non-overtriage). The agent demonstrated almost perfect self-consistency (Fleiss' κ = 0.850 [95% CI: 0.786–0.909], 83% perfect 5/5 agreement). The system operated at a median cost of $0.34 per triage in 94.5 seconds.

**Conclusions.** These findings demonstrate that autonomous AI agents can perform reliable, contextual clinical triage of RPM vital signs with sensitivity to clinical deterioration that, in this retrospective evaluation under our study design, exceeded that of individual human clinicians reviewing pre-assembled context summaries. Sentinel substantially outperformed rule-based thresholds, achieved agreement with clinician panels comparable to inter-clinician agreement, and—in head-to-head leave-one-out comparison—detected emergencies more reliably than any individual clinician while maintaining a clinically defensible overtriage profile. The agent's ability to systematically retrieve and synthesize full patient context for every reading addresses the core limitation that undermined prior RPM trials. At $0.34 per triage, this approach offers a scalable path toward the intensive, contextualized monitoring model that TIM-HF2 showed reduces mortality.

# 1. Introduction

Noncommunicable chronic diseases account for 74% of global deaths.[1] In the U.S., 90% of the nation's $4.9 trillion annual health expenditures are for people with chronic and mental health conditions.[2,3] Remote patient monitoring (RPM)—the collection and clinical review of patient-generated health data outside traditional care settings—has been widely promoted as a solution for managing chronic conditions including heart failure (HF), hypertension, and diabetes.[4] By 2025, over 70 million Americans were projected to use RPM devices. RPM adoption in the United States has surged in recent years, with over 13.5 million Medicare remote monitoring services delivered between 2019 and 2023, generating an unprecedented volume of clinical data.[5,6] Yet the central promise of RPM—that more data enables better care—has been undermined by a fundamental paradox: the data floods that RPM produces overwhelm the very clinicians who must act on them.

## *1.1 The Failed Trials and Their Lessons*

Three landmark randomized controlled trials in heart failure telemonitoring illustrated this paradox with devastating clarity.

Tele-HF deployed telephone-based daily monitoring for HF patients and found no significant improvement in readmission or death.[7] It also revealed the mechanism of failure: the average patient generated 35 variances of alerts requiring staff-patient contact over 6 months, yet there were no specified algorithms connecting signals with responses. Clinicians were buried in threshold-crossing alerts with no way to separate signal from noise.

BEAT-HF combined telemonitoring with health coaching after HF discharge and similarly failed to reduce 180-day readmissions.[8] The investigators were remarkably candid about systemic flaws, acknowledging that "the physiological signals of changes in daily weights and increased symptoms may not provide adequate warning of impending decompensation" without clinical context. Weight—the primary monitored parameter—was a poor surrogate for cardiac filling pressures when interpreted in isolation.

TIM-HF1 also failed to show benefit in the overall population, though a pre-specified subgroup analysis suggested benefit among patients without major depression who had a recent HF hospitalization.[9,10]

A meta-analysis found that the lowest mortality was associated with interventions performed within one day of a vital sign change;[11] later syntheses note persistent heterogeneity in home telemonitoring systems.[12]

Drawing on early neutral trials and the 2023 JACC Scientific Statement,[13] we summarize recurring architectural failure modes:

- *Data flood without intelligent filtering.* Every threshold crossing demanded human attention regardless of clinical significance, producing classic alert fatigue. This systemic vulnerability is well-documented: only 15% were clinically relevant,[14] RPM and CDS-related alert override rates range from 37.9% to 59.6%,[15,16] threshold-based alerting at population-scale RPM produces clinically unsustainable notification volumes,[7,8] and the Joint Commission has identified alarm-related sentinel events including patient deaths.[17]
- *No structured response protocol.* Detection of an abnormal value triggered no specified algorithm for response, creating a "broken loop" or inconsistency from detection to action.
- *Lack of contextual interpretation.* Raw vital signs were assessed without medication context, comorbidity awareness, baseline personalization, or trend analysis.

- *Catastrophic adherence failure.* Systems requiring active patient engagement suffered rapid attrition and data loss.
- *No population targeting.* Monitoring all patients equally, regardless of risk, diluted resource allocation and statistical power.

## *1.2 TIM-HF2: Proof That Structured Monitoring Saves Lives*

TIM-HF2 systematically addressed these failure modes and succeeded.[18] By deploying a 24/7 physician-staffed telemedical center with structured response protocols, TIM-HF2 reduced all-cause mortality by 30% (HR 0.70, 95% CI 0.50–0.96; p=0.028) and unplanned hospitalization days by 20%. Extended follow-up confirmed that benefits disappeared within one year of stopping the intervention.[19] The key element was that a well-structured telemedical center providing 24/7 service was present, with empowered clinicians who could change medication regimens.[18,20] TIM-HF2 proved that responsive, contextualized, structured monitoring saves lives. But it solved the problem by deploying expensive 24/7 human infrastructure that cannot scale.

## *1.3 AI Agents as a Paradigm Shift*

We propose that effective monitoring requires not better algorithms or lower thresholds, but a fundamentally different class of system: an autonomous AI agent capable of clinical reasoning with full patient context and structured action capabilities.

Such a system must be precisely calibrated between two failure modes. *Undertriage*—classifying a deteriorating patient as stable—risks missed emergencies and preventable harm. But *overtriage* may be equally dangerous at scale: if an AI system flags 50% of readings as urgent, as simple threshold rules do, it recreates the exact data flood that overwhelmed clinicians in Tele-HF and BEAT-HF, rendering the system clinically useless regardless of its sensitivity. An effective AI triage agent must therefore demonstrate not only that it catches true emergencies, but that it does so without generating the unsustainable alert volumes that erode clinician trust and attention—the very alarm fatigue that prior RPM trials proved fatal to their own efficacy.

Large language models (LLMs) have achieved high performance on medical QA and clinical knowledge benchmarks.[21-24] However, standalone LLMs lack structured data access for real-world clinical decision-making. The emerging paradigm of agentic AI—LLMs augmented with tool-calling capabilities—addresses this limitation. Model Context Protocol (MCP) provides a standardized interface for connecting AI agents to structured data sources.[25]

We present **Sentinel**, to our knowledge the first deployed autonomous AI agent system using MCP for clinical triage in outpatient RPM. This system directly addresses the five architectural failure modes of prior trials. Its 21 clinical tools provide the deep contextual interpretation that was previously missing, and its AI-driven reasoning acts as an intelligent filter to prevent data floods. Crucially, the agent autonomously determines which clinical data to retrieve at runtime—dynamically selecting from patient history, medications, diagnoses, clinical notes, and prior encounters based on the specific clinical context of each reading—rather than following a fixed data retrieval sequence. Its structured analytical output provides a clear, actionable signal, closing the loop from detection to response. In this study, we evaluated Sentinel's reliability, compared its performance against rule-based baselines and human clinicians, and conducted a leave-one-out analysis comparing the agent against each individual clinician alongside independent clinical adjudication of agent overtriage cases.

## 2. Related Work

### 2.1 Commercial RPM Platforms

Many RPM programs still operationalize alerts with heuristic threshold rules, while some vendors add machine learning (ML)-derived indices.[26] Several platforms have introduced varying degrees of algorithmic sophistication. Biofourmis employs ML-based physiological modeling to generate personalized health indices, demonstrating high sensitivity (94.1%) and specificity (88.9%) for detecting clinical deterioration in a single-center observational study of 34 hospitalized mild COVID-19 patients.[27] The platform, originally known for hospital-at-home and acute care solutions, has since expanded across the care continuum through its 2024 merger with CopilotIQ, creating an end-to-end platform spanning pre-surgical optimization, acute, post-acute, and chronic in-home care.[28] Health Recovery Solutions, recognized as Best-in-KLAS for RPM in 2023, offers nurse-staffed clinical triage services, which provide RN-led monitoring and escalation for RPM patients.[29,30] Masimo SafetyNet provides continuous SpO2 monitoring with configurable alert thresholds based on institutional clinical protocols, primarily relying on clinician-configured parameters rather than automatically adaptive baselines.[31] Across these platforms, a common architectural pattern persists: alerts are generated by fixed or simple statistical rules applied to individual vital sign readings, without integrating the broader clinical context—medications, diagnoses, hospitalization history, or temporal trends—that clinicians use to interpret vital signs. To the best of our knowledge, no commercial platform combines LLM-agent-powered clinical reasoning with dynamic, context-driven data retrieval for outpatient RPM triage.

### 2.2 LLMs and Agentic AI in Clinical Triage

Although large language models (LLMs) demonstrate baseline clinical competence across medical benchmarks,[21-24] their application to patient triage yields mixed results.[32] Single-prompt models achieve physician-level accuracy in emergency department triage,[33] yet exhibit systematic over-triage—a critical calibration risk for deployment in remote patient monitoring.[32] Clinical architectures are consequently shifting toward tool-augmented workflows: COMPOSER-LLM demonstrated that pairing an LLM with a predictive model to autonomously extract clinical context during high-uncertainty cases can substantially reduce false alarms in sepsis detection.[34]
Beyond single-prompt classification, autonomous clinical agents employ multi-step reasoning with structured tool access.[35] However, benchmarks confirm that agentic clinical workflows are substantially harder than static question-answering, with model accuracy often dropping.[36,37] While standardized tool-use frameworks such as MCP are emerging in healthcare,[38] clinical AI agents have remained largely confined to simulated environments.[39-41] To our knowledge, Sentinel is the first LLM agent utilizing MCP tools deployed in a clinical production environment for direct patient care.

# 3. Methods

## *3.1 Study Population*

The main evaluation study population comprised 340 unique patients enrolled in AnsibleHealth's polychronic care program across 25 U.S. states, with 500 RPM device readings drawn from a representative 24-hour period from a weekday in 2026. Patient demographics were extracted from structured EHR data using the same tools available to the agent (Table 1; N=339, with 1 patient excluded due to missing data). Patients were predominantly elderly (mean age 70.3 years) and female (60.2%). The cohort was medically complex, with high rates of hypertension (45.1%), heart failure (31.9%), and diabetes mellitus (31.0%), and a median of 19 documented ICD-10 conditions per patient. Readings comprised 228 blood pressure (45.6%), 195 pulse oximetry (39.0%), and 77 body weight (15.4%) measurements.

**Table 1. Study Population Demographics (N=339)**

| Characteristic | Value |
| --- | --- |
| Patients, N | 339 |
| Age, years: mean (SD) | 70.3 (9.5) |
| Age, years: median (IQR) | 70 (65–77) |
| Age, years: range | 32–91 |
| Age <50 | 10 (2.9%) |
| Age 50–59 | 32 (9.4%) |
| Age 60–69 | 116 (34.2%) |
| Age 70–79 | 122 (36.0%) |
| Age 80–89 | 56 (16.5%) |
| Age ≥90 | 3 (0.9%) |
| Female sex | 204 (60.2%) |
| Male sex | 135 (39.8%) |
| Major Comorbidities | |
| Hypertension | 153 (45.1%) |
| Heart failure | 108 (31.9%) |
| Diabetes mellitus | 105 (31.0%) |
| Coronary artery disease | 92 (27.1%) |
| COPD | 82 (24.2%) |
| Obesity | 77 (22.7%) |
| Chronic kidney disease | 73 (21.5%) |
| Documented conditions (ICD-10), median (range) | 19 (1–399) |
| RPM Device Readings | 500 |
| Blood pressure cuff | 228 (45.6%) |
| Pulse oximeter | 195 (39.0%) |
| Body weight scale | 77 (15.4%) |

*Note: Demographics sourced via the same tools used by the agent. One patient was excluded due to missing EHR data.*

## 3.2 System Architecture

We conducted a retrospective evaluation of Sentinel using data from AnsibleHealth, a value-based remote patient monitoring organization serving patients with chronic conditions including heart failure, hypertension, COPD, and diabetes across multiple health system partnerships. The evaluation dataset comprised vital sign readings from three home device types: blood pressure cuffs (systolic/diastolic blood pressure, pulse rate), pulse oximeters (SpO2, pulse rate), and weight scales (bodyweight). All analyses were performed offline on historical data; no patient care decisions were made during the study.

### 3.2.1 AI Triage Agent

The triage agent used Anthropic's claude-opus-4-6 model operating via the Claude Agent SDK in an agentic framework with structured tool access.[42] All inference parameters were set to Anthropic's defaults (temperature = 1.0, no top_p override); no deterministic seeding was applied, as the study design intentionally leveraged stochastic variability to assess reliability. Unlike single-prompt classification systems, the agent autonomously determined which clinical data to retrieve, how many reasoning steps to take, and what contextual factors to weigh–executing a variable-length, multi-step clinical reasoning workflow for each reading.

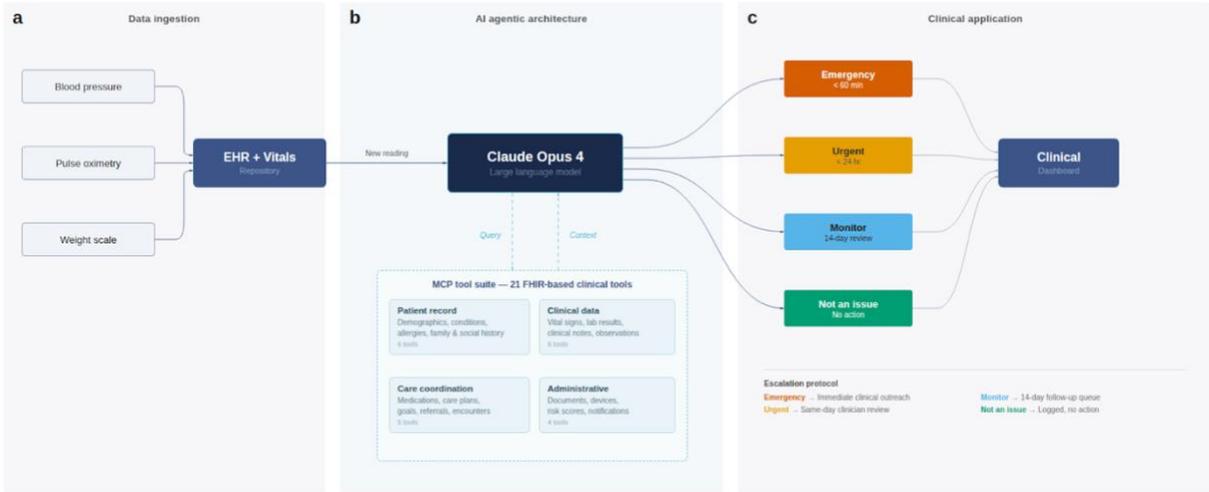

**Figure 1. Sentinel System Architecture**

The agent framework provided access to 21 tools organized across three services (Table 2). The patient clinical data service exposed 17 tools for retrieving structured patient data via Model Context Protocol. The ICD-10 terminology service provided diagnostic code lookup. Three native system tools (Read, Bash, Grep) enabled additional data manipulation when needed. Through these tools, the agent had access to the comprehensive electronic health record (EHR), including data aggregated from health information exchanges (HIEs) across the patient's care network. Available data sources included patient demographics, ICD-10-coded active diagnoses, current medications with dosing, vital sign history with baseline statistics, recent encounters including hospitalizations, clinical documentation from both internal and external providers, clinician annotations on prior readings, outreach activity and call records, HIE-sourced care summaries, program enrollment status, and equipment records.

**Table 2. MCP Tool Inventory (N=500 trials)**

| Tool | Calls | Trials Used | % Trials |
|---|---|---|---|
| *Core tools (100% usage):* | | | |
| getpatientdemographics | 500 | 500 | 100.0% |
| get_conditions | 500 | 500 | 100.0% |
| get_medications | 500 | 500 | 100.0% |
| get_encounters | 500 | 500 | 100.0% |
| getvitalhistory | 500 | 500 | 100.0% |
| *Frequent tools (>25% usage):* | | | |
| getrecentactivity | 471 | 462 | 92.4% |
| search_notes | 356 | 308 | 61.6% |
| getprogramstatus | 309 | 300 | 60.0% |
| getclinicalnotes | 288 | 270 | 54.0% |
| getcalldetail | 450 | 249 | 49.8% |
| gethiesummary | 146 | 145 | 29.0% |
| *Occasional tools (<25% usage):* | | | |
| getlatestvitals | 129 | 122 | 24.4% |
| Read (native) | 153 | 58 | 11.6% |
| Grep (native) | 90 | 56 | 11.2% |
| Bash (native) | 92 | 53 | 10.6% |
| getsessionhistory | 5 | 5 | 100.0% |
| getpatientequipment | 3 | 3 | 100.0% |
| icd10lookupcode | 8 | 2 | 25.0% |
| getexercisevitals_trend | 1 | 1 | 100.0% |
| getattendancesummary | 1 | 1 | 100.0% |
| get_observations | 1 | 1 | 100.0% |

The agent selectively invoked tools based on clinical context rather than following a fixed retrieval sequence. Five core tools were called on every trial (get_patient_demographics, get_conditions, get_medications, get_encounters, get_vital_history), while additional tools were called situationally. The mean number of tool calls per trial was 10.1 (range 6–24). Higher-acuity classifications correlated with deeper investigation: emergency cases averaged 14.2 tool calls versus 8.7 for non-issue cases, a 63% increase.

The agent assigned one of four severity levels: EMERGENCY (clinical outreach within 60 minutes), URGENT (within 24 hours), MONITOR (review within 14 days), and NOT AN ISSUE (no action required) (Figure 2). The system prompt included explicit escalation guardrails instructing the agent not to escalate solely on the basis of absent recent clinical contact, longstanding uncontrolled chronic disease, or high comorbidity burden without evidence of acute instability. Independent of severity, the agent classified the required follow-up into one of six action types (no action, equipment resolution, patient education, clinical review, urgent review, or care coordination) to facilitate clinical routing.

**EMERGENCY** — *Clinical outreach within 60 minutes*

Reading indicates immediate safety risk where delay could materially increase harm. Requires emergent clinician action.

**Typical triggers:**
- Rapid, severe deterioration pattern across recent measurements
- Credible reading incompatible with safety if delayed
- Hemodynamic instability (e.g., >50 mmHg systolic swing within hours)
- Rapidly worsening trajectory approaching critical limits

**URGENT** — *Clinical outreach within 24 hours*

Clinically meaningful abnormality requiring same-day clinician assessment, or worsening trend that is concerning but not an immediate crisis.

**Typical triggers:**
- Clear abnormality requiring same-day clinician review or intervention
- Worsening trend that is concerning but not immediately dangerous
- New or rapidly worsening pattern over hours to 1–2 days
- Dangerously high/low vitals with relevant comorbidities

**MONITOR** — *Review within 14 days*

Concerning but not immediately dangerous, OR persistent equipment/data problem needing non-urgent follow-up.

**Typical triggers:**
- Gradual weight gain in CHF; slowly rising BP in renal disease
- Trending SpO$_2$ decline without acute distress
- Device producing impossible values (e.g., scale reading 1000+ kg)
- Household member repeatedly using patient's device
- Clinician has already reviewed this pattern per recent notes

**NOT AN ISSUE** — *No action required*

After full context review, the reading is non-actionable and requires no clinical or operational follow-up of any kind.

**Typical triggers:**
- Single device glitch that self-resolved
- Single outlier in an otherwise stable patient
- Reading within expected range for this patient's known baseline

**Tie-Breaker Rules**

| | |
|---|---|
| **EMERGENCY vs. URGENT:** | If immediate safety risk is uncertain, classify as URGENT. |
| **URGENT vs. MONITOR:** | If immediate clinical danger is uncertain, choose MONITOR. |
| **MONITOR vs. NOT AN ISSUE:** | If uncertain, choose MONITOR. Better to over-monitor than miss a developing problem. |

*Escalation guardrail: Do NOT escalate solely based on absent recent clinical contact, longstanding uncontrolled chronic disease, or high comorbidity burden — without evidence of acute instability.*

**Figure 2. Severity Classification Decision Guidelines.** Definitions, response timeframes, clinical examples, and tie-breaker rules provided to all raters (AI agent and human clinicians). Escalation guardrails prevented classification based solely on absent clinical contact, chronic disease burden, or comorbidity without evidence of acute instability.

### 3.2.2 Research Configuration

For all research trials, every tool call enforced temporal correctness by including a parameter set to the reading's original timestamp, ensuring the agent accessed only data available at the time the reading was taken. The agent received only raw vital sign values, patient identifiers, and reading metadata; it did not receive prior data quality-control filter results or prior trial outputs. Each trial was fully independent with no cross-trial memory. The agent operated with a maximum of 15 reasoning turns per trial and a 120-second timeout.

## 3.3 Study 1: Agent Inter-Rater Reliability

To assess the stability of the AI agent triage, we conducted a multi-run inter-rater reliability study treating each independent execution of the AI agent as a separate "rater." One hundred random vital sign device readings were drawn from AnsibleHealth's production database, comprising 34 blood pressure cuff readings, 33 pulse oximeter readings, and 33 body weight readings, spanning 92 unique patients. Each device reading was independently triaged 5 times by separate AI agents, yielding 500 total trials. The agent used the 4-level classification scheme (EMERGENCY/URGENT/MONITOR/NOT AN ISSUE) as defined in Section 3.2.1.

The primary outcome was Fleiss' kappa coefficient for the four-level classification across 5 raters, with 95% confidence intervals computed via 2,000 bootstrap resamples. Because the four triage levels are ordinal, we also calculated linearly weighted Fleiss' kappa to account for the severity of disagreements. Secondary outcomes included the perfect agreement rate (all 5 runs identical), per-vital-type agreement, and operational metrics.

## 3.4 Study 2: Agent vs. Rule-Based Baselines

We compared the agent's triage decisions against two rule-based baseline systems on the same 500 readings: a (1) Fixed Threshold Baseline, and an (2) Adaptive Baseline. The Fixed Threshold Baseline combined clinical guideline cutoffs,[43] heart failure rules-of-thumb,[44] and persistence rules,[45] producing a 2-level output (URGENT/NOT AN ISSUE). As a secondary fixed comparator, we included a modified National Early Warning Score 2 (NEWS2),[46] adapted for the limited parameters available from home RPM devices, producing a 4-level classification (EMERGENCY/URGENT/MONITOR/NOT AN ISSUE). The Adaptive Baseline used a rolling 30-day z-score detector with patient-specific statistical bounds to generate a 4-level classification (EMERGENCY/URGENT/MONITOR/NOT AN ISSUE). The primary outcome was alert volume (rate), and the secondary outcomes were agreement between agent and baseline systems. Full baseline specifications and the threshold search strategy are provided in the Appendix.

## 3.5 Study 3: Agent vs. Human Clinician Validation

### 3.5.1 Study Design

Six clinical reviewers (3 physicians: MD1, MD2, MD3; 3 nurse practitioners: NP1, NP2, NP3) independently graded the same 500 RPM readings using the 4-level classification scheme (EMERGENCY/URGENT/MONITOR/NOT AN ISSUE). The study used a connected matrix design: each sample was independently assigned to 3 of the 6 reviewers, and each reviewer graded 330 total readings (250 unique samples plus 20 anchor samples presented multiple times for intra-rater reliability assessment). Reviewers were blinded to each other's annotations and to all automated algorithm decisions.

Reviewers accessed a structured clinical interface displaying the vital sign reading alongside comprehensive patient context: interactive 30-day vital sign trend charts, the patient's active problem list

(ICD-10-coded conditions), hospitalization history, clinical notes from the preceding 12 months, and recent patient call summaries. Reviewers had unlimited time for each assessment.

A key methodological challenge in evaluating any AI triage system is the absence of a definitive reference standard. Unlike diagnostic tests with objective ground truth (e.g., biopsy-confirmed pathology), clinical triage severity is inherently subjective—reasonable clinicians routinely disagree on the appropriate level of escalation for identical patient presentations, particularly in RPM triage with its limited information. Clinical follow-up results of the triage would provide some confirmation, albeit still with subjectivity. However, such follow-up of isolated RPM vitals is impracticable retrospectively. To rigorously validate Sentinel's performance despite this "reference standard problem," we employed a three-pronged strategy. First, we quantified human inter-reviewer agreement and intra-rater reliability (Section 3.5.2) to establish the ceiling and variability of human performance, contextualizing agent-human disagreements within the range of clinician-to-clinician disagreement. Second, we conducted a leave-one-out (LOO) analysis (Section 3.5.3) that constructs an independent reference standard for each reviewer—human and AI alike–so the agent is evaluated using the same methodology applied to each individual clinician rather than being compared against a consensus it had no role in forming. Third, all cases where the agent severely escalated beyond the human majority underwent independent clinical adjudication (Section 3.5.4) to determine whether these disagreements represented algorithmic error or clinically defensible escalation. Together, these three validation layers pressure-test the agent's results from complementary angles—demonstrating the degree of human subjectivity, enabling fair head-to-head comparison, and directly interrogating the clinical validity of every agent disagreement.

### 3.5.2 Endpoints

The primary reference standard was established using a majority vote among the three assigned reviewers for each case. Primary endpoints included agreement between the agent and the human majority vote (both exact 4-class and binary [alert vs. not alert]), sensitivity (recall), precision (PPV), specificity, negative predictive value (NPV), false positive rate (FPR), and false negative rate (FNR). Disagreements were further analyzed to quantify the frequency and direction of any agent undertriage or overtriage (e.g., agent: monitor; human: emergency).

Secondary endpoints included human inter-reviewer agreement, intra-rater reliability, and triage duration to contextualize the agent's performance against the inherent subjectivity of the clinical task.

### 3.5.3 Leave-One-Out Individual Clinician Analysis

The majority-vote reference standard, while conventional, introduces a structural asymmetry: the agent is compared against a consensus of three clinicians, but individual clinicians are evaluated as part of that same consensus. A clinician who systematically undertriages may still achieve high "agreement" with the majority if the other two reviewers happen to align with their assessment. The leave-one-out analysis corrects this asymmetry by constructing an independent reference standard for each clinician: for a given sample, the LOO majority is formed from the two reviewers who did not include the clinician being evaluated. This places every rater—human and AI alike—on equal methodological footing, each judged against a reference standard they did not help create. Samples where the remaining two reviewers disagreed (no majority exists) were excluded from that clinician's LOO evaluation. The agent was then evaluated against the identical LOO reference standards, enabling direct head-to-head comparison of individual clinician performance versus the agent on the same cases with the same evaluation criteria.

This design is critical because it reveals whether the agent's sensitivity and specificity are genuinely superior to individual clinicians or merely an artifact of comparison against a pooled consensus that smooths over individual variation.

Across the six clinicians, exclusion due to reviewer disagreement ranged from 35.6% to 46.0% of each clinician's 250 assigned samples (LOO evaluable N per clinician: 135–161), reflecting substantial baseline disagreement among human reviewers on ambiguous cases. The agent was evaluated against the same LOO subsets.

### 3.5.4 Clinical Adjudication of Agent Overtriage Cases

Of the 105 agent overtriage cases (where the agent rated higher than the human majority by ≥1 severity level), we prioritized the 17 most severe disagreements—cases where the agent exceeded the majority by ≥2 severity levels—for independent human expert adjudication. Single-level disagreements (N=88) fall within the demonstrated range of inter-clinician variability (clinician exact agreement 59.7%, Section 4.3) and represent adjacent severity categories where reasonable clinicians routinely disagree. Human expert adjudication was therefore reserved for cases where, if the agent were wrong, the clinical consequences would be most significant.

Two independent physician reviewers, blinded to each other's assessments, independently re-graded each of the 17 cases using the same 4-level classification scheme (EMERGENCY/URGENT/MONITOR/NOT AN ISSUE). Both reviewers had access to the vital sign reading, interactive 30-day vital sign trend charts, the patient's active problem list, clinical notes, the agent's full reasoning with cited data sources, and the original reviewer classifications. Each case was then classified into one of three categories by comparing the adjudicator's re-grade to both the agent's classification and the original human majority vote: JUSTIFIED (adjudicator matched or exceeded the agent's severity level, indicating the agent's escalation was clinically appropriate), DEBATABLE (adjudicator upgraded from the original majority but assigned a severity level below the agent, indicating the majority was too conservative but the agent may have over-escalated), or TRUE OVERTRIAGE (adjudicator agreed with the original majority vote, confirming the agent's escalation was not clinically warranted).

As a complementary analysis covering all 105 overtriage cases, we performed an automated disagreement characterization using two independent large language models (Claude Opus 4.6 and Gemini 3.1 Pro) to systematically categorize the clinical phenotypes underlying the agent's overtriage. Rather than rendering binary justified/unjustified verdicts, this automated analysis focused on identifying the recurring clinical patterns that drove agent-human disagreement.

## 4. Results

### 4.1 Agent Self-Consistency (Study 1)

The agent achieved a Fleiss' kappa of 0.850 (95% CI: 0.786–0.909) for the four-level classification across 5 independent runs on 100 samples, indicating almost perfect agreement by Landis-Koch criteria.[47] All confidence intervals were computed via 2,000 bootstrap resamples (resample size=100 readings, with replacement). Perfect five-way agreement (identical classification across all 5 runs) was achieved on 83 of 100 samples (83%).

**Table 3. Agent Self-Agreement: Fleiss' Kappa (N=100, 5 runs).** Overall Fleiss' κ (4-level): 0.850. Perfect 5/5 agreement: 83%.

| Category | Fleiss' κ (95% CI) | Distribution (% of 500 runs) |
| --- | --- | --- |
| Emergency | 0.798 (0.370–1.000) | 18 (3.6%) |
| Urgent | 0.621 (0.326–0.811) | 34 (6.8%) |
| Monitor | 0.842 (0.761–0.913) | 166 (33.2%) |
| Not an issue | 0.923 (0.868–0.971) | 282 (56.4%) |
| Overall | 0.850 (0.786–0.909) | 500 (100%) |

## 4.2 Rule-Based Baseline Comparison (Study 2)

On the 500-reading subset, the fixed threshold baseline flagged 55.8% of readings as urgent–a rate 1.7 times higher than the agent's combined urgent and emergency rate of 33.2% (Table 4). Notably, the agent classifies 12.6% of readings as emergency (vs. the fixed threshold's 0%) and 20.6% as urgent, reflecting its sensitivity to high-acuity readings that rule-based systems miss entirely. The adaptive baseline was the most conservative, flagging only 9.4% as urgent or emergency. The agent used its four-level classification to differentiate clinically, with 20.2% classified as monitor—readings warranting review but not same-day action.

**Table 4. Alert Rate Comparison: Fixed Threshold vs. Adaptive vs. Agent (N=500)**

| System | URGENT/EMERGENCY | MONITOR | NOT AN ISSUE |
| --- | --- | --- | --- |
| Fixed threshold | 279 (55.8%) | 57 (11.4%) | 164 (32.8%) |
| Adaptive | 47 (9.4%) | 63 (12.6%) | 390 (78.0%) |
| Agent | 166 (33.2%) | 101 (20.2%) | 233 (46.6%) |

*Note: On the same 500 readings, the fixed threshold classified 55.8% as urgent/emergency compared to the agent's 33.2%, while the adaptive baseline classified only 9.4% as urgent/emergency. These alert rate differences are contextualized in Section 4.4.2 against the human majority-vote reference standard.*

## 4.3 Human Clinician Performance and Reliability (Study 3)

To contextualize the agent's performance, we first assessed the reliability of the human reviewers.

### 4.3.1 Intra-Rater Reliability

Each reviewer re-graded 20 of their own samples. Pairwise exact agreement averaged 75.8% (range: 55%–95%).

**Table 5. Intra-Rater Reliability.** 20 unique samples per reviewer, repeated 5 times.

| Reviewer | Exact Match | Percentage |
| --- | --- | --- |
| MD1 | 11/20 | 55.0% |
| MD2 | 18/20 | 90.0% |
| MD3 | 17/20 | 85.0% |
| NP1 | 12/20 | 60.0% |
| NP2 | 14/20 | 70.0% |
| NP3 | 19/20 | 95.0% |
| Mean | 15.2/20 | 75.8% |

By comparison, the agent achieved 83% perfect five-way agreement in Study 1–exceeding the self-consistency of most individual human reviewers.

*4.3.2 Inter-Reviewer Agreement*

Pairwise exact agreement between any two clinicians on the same sample averaged 59.7% (range: 46.2%–69.1%), highlighting the inherent subjectivity of RPM triage (Figure 3). Reviewer consensus (all 3 reviewers unanimous) was achieved on 42.8% (214/500) of samples (Table 6). Three-way splits (all 3 reviewers selecting different triage levels) occurred on 6.6% (33/500) of samples.

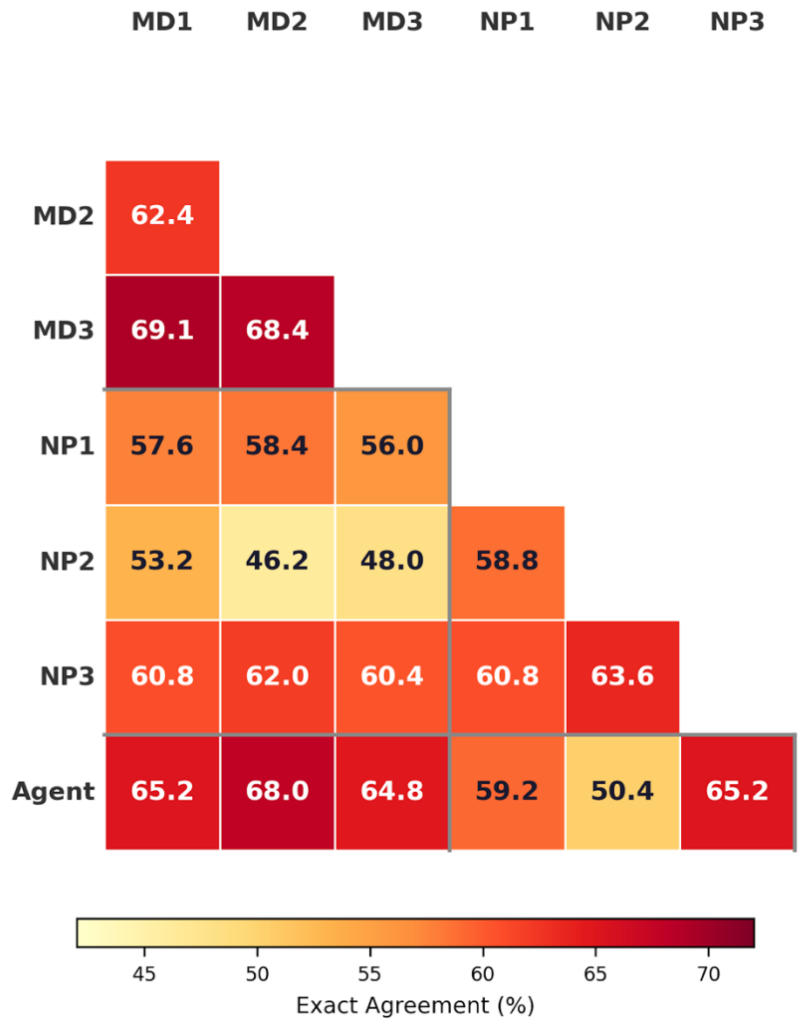

**Figure 3. Inter-Reviewer Pairwise Exact Agreement Matrix.** Average agreement of 15 inter-reviewer pairs was 59.7% (range: 46.2%–69.1%), and the average agreement of agent-reviewer pairs was 62.1% (range: 50.4%–68.0%).

**Table 6. Inter-Reviewer Agreement Distribution**

| Metric | Value |
| --- | --- |
| Mean pairwise exact agreement | 59.7% (range: 46.2%–69.1%) |
| Unanimous consensus (3/3) | 214/500 (42.8%) |
| Majority consensus (≥2/3) | 467/500 (93.4%) |
| Three-way split | 33/500 (6.6%) |

### 4.3.3 Reviewer vs. Agent Agreement

Pairwise agreement between individual reviewers and the agent averaged 62.1%, ranging from 50.4% to 68.0% (Table 7). This exceeded the average inter-reviewer agreement of 59.7% by 2.4 percentage points, indicating that humans agreed with the agent slightly more than they agreed with each other. Binary agreement (actionable vs. non-actionable) averaged 83.5%.

**Table 7. Reviewer-Agent Pairwise Agreement**

| Reviewer | Exact Agreement | Binary Agreement | Within ±1 |
|---|---|---|---|
| MD1 | 163/250 (65.2%) | 210/250 (84.0%) | 236/250 (94.4%) |
| MD2 | 170/250 (68.0%) | 211/250 (84.4%) | 236/250 (94.4%) |
| MD3 | 162/250 (64.8%) | 212/250 (84.8%) | 235/250 (94.0%) |
| NP1 | 148/250 (59.2%) | 206/250 (82.4%) | 238/250 (95.2%) |
| NP2 | 126/250 (50.4%) | 210/250 (84.0%) | 241/250 (96.4%) |
| NP3 | 163/250 (65.2%) | 204/250 (81.6%) | 232/250 (92.8%) |
| Average | 62.1% | 83.5% | 94.5% |

*4.4 Agent vs. Human Majority Vote Reference Standard*

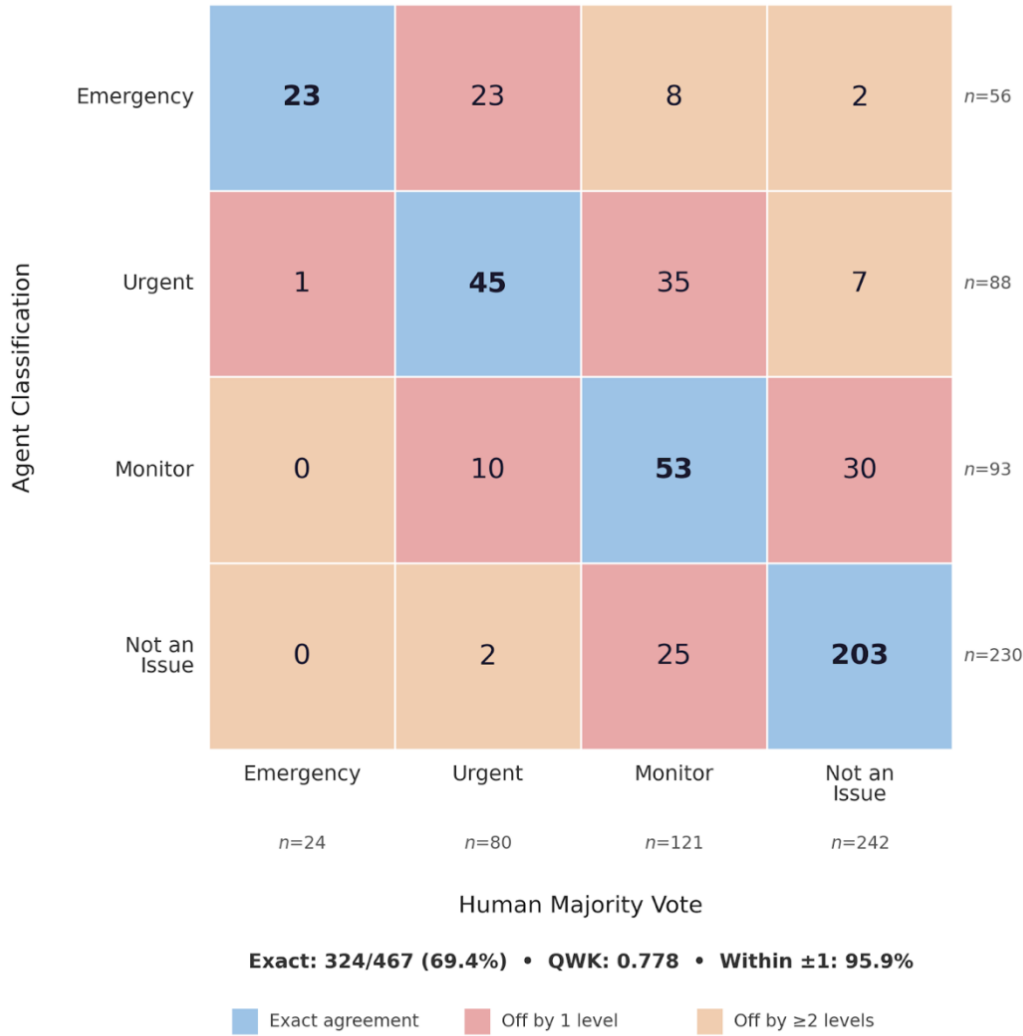

**Figure 4. Agent vs. Human Majority Vote Confusion Matrix (N=467)**

Thirty-three samples with three-way splits among reviewers (no majority vote) were excluded from accuracy analyses, yielding an evaluable set of 467 samples. Table 8 summarizes agent performance against the majority-vote reference standard; these metrics should be interpreted with the caveat that the reference standard itself carried substantial uncertainty (clinician unanimity on only 42.8% of samples; see Section 4.3.2), and that the agent or a minority reviewer may have been correct in cases scored as disagreements. Overall, four-level exact accuracy was 69.4% (324/467), with a quadratic-weighted kappa of 0.778 (substantial agreement; 95% CI: 0.728–0.819) and 95.9% of classifications falling within one severity level of the reference standard.

Per-category recall (Table 9) demonstrated strong detection of high-acuity readings: emergency sensitivity reached 95.8% (23/24) and urgent recall was 56.3% (45/80). The agent's disagreement profile favored overtriage over undertriage: agent overestimation of severity (relative to human majority) occurred in 105 cases (22.5%) versus underestimation in 38 cases (8.1%), a ratio of 2.8:1 favoring clinical caution.

**Table 8. Agent Performance vs. Majority Vote Reference Standard (N=467).** Thirty-three samples with three-way splits among reviewers (no majority vote) were excluded from analysis.

| Metric | Value |
| --- | --- |
| Four-level exact accuracy | 324/467 (69.4%) |
| Quadratic-weighted kappa | 0.778 (95% CI: 0.728–0.819) |
| Within ±1 severity level | 448/467 (95.9%) |
| Overtriage cases | 105/467 (22.5%) |
| Undertriage cases | 38/467 (8.1%) |

**Table 9. Agent Per-Category Performance vs. Majority-Vote Reference Standard (N=467)**

| Category | Reference Standard Prevalence | Recall | Precision |
| --- | --- | --- | --- |
| EMERGENCY | 24 (5.1%) | 23/24 (95.8%) | 23/56 (41.1%) |
| URGENT | 80 (17.1%) | 45/80 (56.3%) | 45/88 (51.1%) |
| MONITOR | 121 (25.9%) | 53/121 (43.8%) | 53/93 (57.0%) |
| NOT AN ISSUE | 242 (51.8%) | 203/242 (83.9%) | 203/230 (88.3%) |

Under the clinically critical binary threshold (Actionable: EMERGENCY+URGENT vs. Non-Actionable: MONITOR+NOT AN ISSUE), the agent achieved 88.5% sensitivity (92/104), 85.7% specificity (311/363), 63.9% PPV (92/144), and 96.3% NPV (311/323). The high NPV indicates that when the agent classifies a reading as non-actionable, it is correct 96.3% of the time (Table 10).

**Table 10. Binary Classification Performance: Actionable vs. Non-Actionable (N=467)**

| Metric | Actionable (E+U vs. M+NI) |
| --- | --- |
| Accuracy | 403/467 (86.3%) |
| Sensitivity | 92/104 (88.5%) |
| Specificity | 311/363 (85.7%) |
| Positive Predictive Value (PPV) | 92/144 (63.9%) |
| Negative Predictive Value (NPV) | 311/323 (96.3%) |

When evaluated against the max-severity reference standard (the highest severity assigned by any of the 3 reviewers, all 500 samples), the agent achieved 80.0% actionable sensitivity and 92.0% specificity, demonstrating robustness even against the most conservative human benchmark.

*4.4.1 Comparison of Agent vs. Rule-Based Baselines Against Human Majority Vote Reference Standard*

To contextualize the agent's performance, we evaluated the same 467 samples against the two rule-based baselines described in Study 2 (fixed threshold with modified NEWS2 and adaptive statistical baseline). Table 11 presents a head-to-head comparison on the key metrics; full per-system confusion matrices are provided in Appendix Tables A1 and A2.

**Table 11. Head-to-Head Performance: Agent vs. Rule-Based Baselines Against Majority-Vote Reference Standard (N=467)**

| Metric | Agent | Fixed Threshold | Adaptive |
| --- | --- | --- | --- |
| Four-level exact accuracy | 324/467 (69.4%) | 250/467 (53.5%) | 234/467 (50.1%) |
| Actionable sensitivity (E+U) | 92/104 (88.5%) | 102/104 (98.1%) | 19/104 (18.3%) |
| Specificity (M+NI) | 311/363 (85.7%) | 215/363 (59.2%) | 341/363 (93.9%) |
| PPV | 92/144 (63.9%) | 102/250 (40.8%) | 19/41 (46.3%) |
| NPV | 311/323 (96.3%) | 215/217 (99.1%) | 341/426 (80.0%) |
| QWK | 0.778 (95% CI: 0.728–0.819) | 0.573 (95% CI: 0.506–0.614) | 0.235 (95% CI: 0.218–0.385) |

The three systems exhibited strikingly different operating points on the sensitivity-specificity trade-off. The agent achieved the best balance: 88.5% actionable sensitivity with 85.7% specificity (QWK 0.778 [95% CI: 0.728–0.819]), compared to the fixed threshold's 98.1% sensitivity at only 59.2% specificity (QWK 0.573 [95% CI: 0.506–0.614]) and the adaptive baseline's 93.9% specificity at only 18.3% sensitivity (QWK 0.235 [95% CI: 0.218–0.385]). The agent's NPV (96.3%) approached the fixed threshold's (99.1%), indicating that both systems rarely missed actionable cases, but the agent accomplished this without the fixed threshold's overwhelming false-positive burden.

The fixed threshold baseline achieved the highest actionable sensitivity (98.1%) but at the cost of flagging 53.5% of all readings as urgent—generating 2.8 times as many false positives as the agent (148 vs. 52)—and only 40.8% PPV. The adaptive baseline showed the opposite failure mode: although its specificity (93.9%) exceeded the agent's, its actionable sensitivity was only 18.3%, missing over 80% of clinician-identified actionable cases—a result of its reliance on patient-specific rolling statistics that normalized chronically abnormal values as within-range.

## 4.5 Individual Clinician vs. Agent Performance (Leave-One-Out Analysis)

To move beyond aggregate majority-vote comparison, we conducted a leave-one-out (LOO) analysis comparing each individual clinician's performance against the agent on identical evaluation criteria. Table 12 presents each clinician's performance against the two-reviewer reference across all of their assigned samples, as well as the pool of all 6 clinicians' subsets to show agent's performance against the two-reviewer reference. The full per-clinician details of agent's LOO performance are included in Appendix Table A4.

**Table 12. Leave-One-Out Analysis: Individual Clinician vs. Agent Performance.** Agent's LOO performance metrics are presented for an aggregated pool of all 6 clinicians' assigned samples.

| Reviewer | LOO Samples | Exact Match | Emergency Sensitivity | Actionable Sensitivity | Overtriage Rate |
|---|---|---|---|---|---|
| MD1 | 154 | 74.0% | 7/9 (78%) | 25/30 (83%) | 16.2% |
| MD2 | 135 | 79.3% | 4/6 (67%) | 14/22 (64%) | 5.2% |
| MD3 | 153 | 78.4% | 4/5 (80%) | 16/26 (62%) | 10.5% |
| NP1 | 161 | 66.5% | 4/6 (67%) | 16/30 (53%) | 20.5% |
| NP2 | 149 | 58.4% | 1/8 (13%) | 20/21 (95%) | 33.6% |
| NP3 | 143 | 74.8% | 4/6 (67%) | 16/25 (64%) | 10.5% |
| Agent | 895 (agg.) | 75.5% | 39/40 (97.5%) | 140/154 (90.9%) | 18.9% |

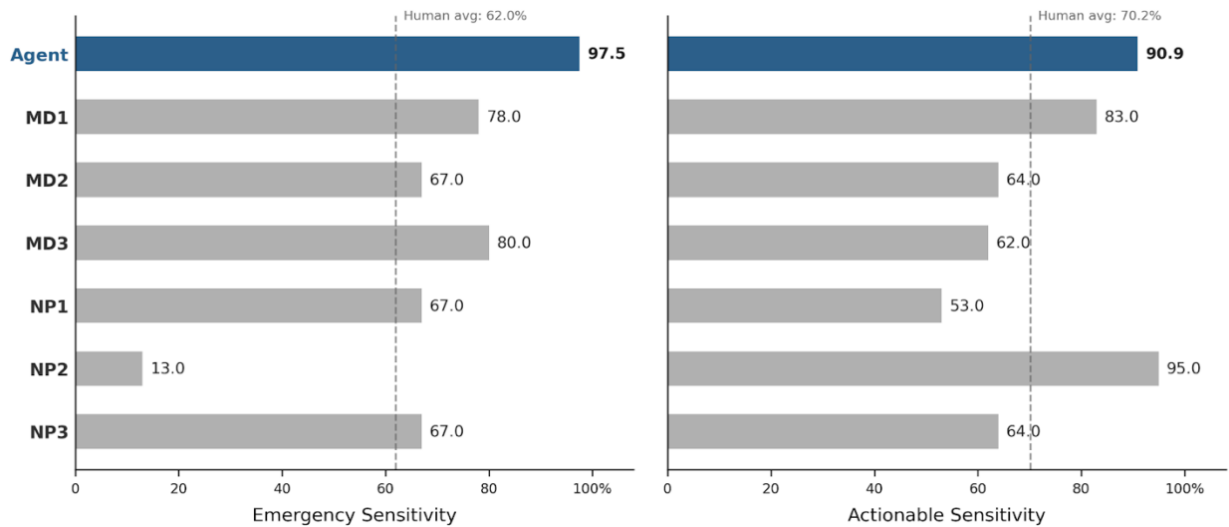

**Figure 5. Leave-One-Out Analysis: Individual Clinician vs. Agent Performance**

The LOO analysis revealed that the agent outperformed every individual clinician on emergency sensitivity (97.5% vs. clinician range 13%–80%, aggregate 60.0%) and actionable sensitivity (90.9% vs. clinician range 53%–95%, aggregate 69.5%). The agent's exact match rate (75.5%) was comparable to most clinicians (range 58.4%–79.3%). Emergency sensitivity showed the most striking contrast: the best individual clinician detected 80% of emergencies, while the agent detected 97.5%. Only one clinician (NP2) achieved higher actionable sensitivity (95%), but at the cost of a 33.6% overtriage rate—the highest of any rater—and the lowest emergency sensitivity (13%).

This analysis demonstrates that the agent functions as a safety net that exceeds any individual clinician's vigilance for critical deterioration. The agent's overtriage rate (18.9%) was moderate—lower than NP1 (20.5%) and NP2 (33.6%)—suggesting that its elevated sensitivity does not come at an unacceptable cost in false escalations.

## 4.6 Clinical Adjudication of Agent Overtriage Cases

Of 467 evaluable readings, agent-human disagreements skewed toward overtriage (105 cases, 22.5%) rather than undertriage (38 cases, 8.1%)—a 2.8:1 ratio favoring clinical caution. To determine whether the agent's most severe overtriage cases represented genuine algorithmic error or clinically defensible escalation, two independent physician reviewers re-graded the 17 cases where the agent exceeded the human majority by ≥2 severity levels (Section 3.5.4). Results are presented in Table 13.

**Table 13. Human Expert Adjudication of Severe Agent Overtriage Cases (Agent–Majority Gap ≥2 Levels; N=17)**

| Category | Expert 1 | Expert 2 | Final Decision |
|---|---|---|---|
| JUSTIFIED (adjudicator matched or exceeded agent) | 7 (41%) | 11 (65%) | 11 (65%) |
| DEBATABLE (adjudicator upgraded from majority, below agent) | 8 (47%) | 5 (29%) | 6 (35%) |
| TRUE OVERTRIAGE (adjudicator agreed with human majority) | 2 (12%) | 1 (6%) | 0 (0%) |
| Total non-overtriage (JUSTIFIED + DEBATABLE) | 15 (88%) | 16 (94%) | 17 (100%) |

Both adjudicators independently validated the agent's escalation in 88–94% of the severe disagreement cases, confirming that the agent's clinical judgement was defensible in the vast majority of instances. Notably, neither adjudicator graded any of these 17 cases as "not an issue"—despite the original majority voting "not an issue" on 9 of the 17 cases (53%). True overtriage, where the adjudicator disagreed with the agent's escalation, was rare: 2 of 17 cases (12%) for Reviewer 1 and 1 of 17 cases (6%) for Reviewer 2. Following independent review, the two adjudicators jointly resolved all 17 cases through consensus discussion, classifying 11 (65%) as JUSTIFIED, 6 (35%) as DEBATABLE, and 0 (0%) as TRUE OVERTRIAGE—yielding a 100% non-overtriage rate and confirming that none of the agent's most severe escalations represented genuine algorithmic error.

As a complementary analysis of all 105 overtriage cases, automated disagreement characterization by two independent large language models identified five recurring clinical phenotypes underlying the agent's escalations: (1) hemodynamic lability (25 cases)—large systolic swings flagged as dangerous even when the current absolute value was benign; (2) trend-based escalation (20 cases)—multi-day worsening trajectories invisible in a single reading; (3) SpO$_2$ on supplemental O$_2$ (15 cases)—the agent correctly noted that desaturation while on supplemental oxygen carries different clinical significance than the same SpO$_2$ on room air; (4) post-discharge vulnerability (12 cases)—escalation based on readmission risk in recently discharged patients; and (5) clinical oversight gaps (~15 cases)—escalation when no clinical contact had occurred for weeks in medically complex patients. These phenotypes describe what the agent's overtriage looks like clinically, providing actionable categories for system refinement.

## *4.7 Operational Metrics*

Table 14 compares operational cost and throughput across the two agent studies. The agent's per-triage cost ranged from $0.32 to $0.34, with zero failures across all 1,000 trials. Figure 6 illustrates a resource gradient of 500-trial run (Study 2, 3) by alert severity—the agent invested progressively more reasoning effort (more tool calls, longer processing, higher cost) for higher-acuity readings, consistent with appropriate clinical diligence rather than uniform processing.

**Table 14. Operational Metrics by Study**

| Metric | Study 1: Agent IRR | Study 2,3: Agent vs. Human |
|---|---|---|
| Trials | 100 x 5 runs | 500 x 1 run |
| Cost per trial (mean) | $0.32 | $0.34 |
| Total cost | $157.89 | $172.36 |
| Duration (median) | 98.7 sec | 94.5 sec (IQR 71.4–131.4) |
| Agent turns (mean) | 11.5 | 12.5 |
| Output tokens (mean) | 4,061 | 3,956 |
| Failure rate | 0/500 (0.0%) | 0/500 (0.0%) |

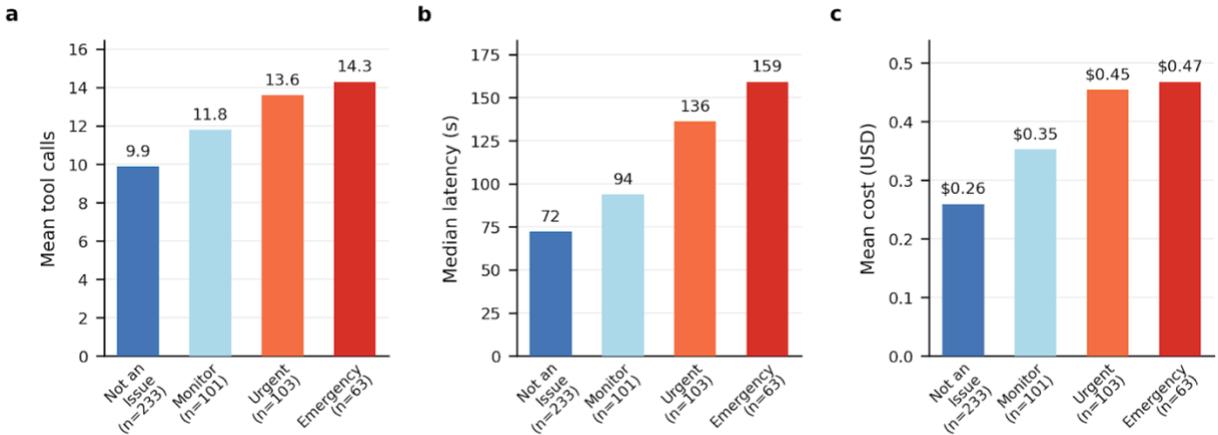

**Figure 6. Tool Usage Gradient by Alert Severity**

## 4.8 Duration of Triage

The agent completed triage in a median of 94.5 seconds (IQR 71.4–131.4 s) per reading (Table 15). Human clinician reviewers completed triage in a median of 7.0–28.0 seconds per reading, depending on the reviewer. These durations were not directly comparable: the human review task was limited to assigning a four-level triage classification given a pre-assembled data summary that included the patient's vital sign history, active conditions, medications, and recent encounters, whereas the agent independently retrieved and synthesized clinical data from 21 MCP tools before rendering its assessment. The human durations therefore excluded the chart review and data gathering time that would be required in a real clinical workflow.

**Table 15. Duration of Triage: Agent vs. Human Reviewers**

| Reviewer | Median (s) | IQR (s) |
|---|---|---|
| Sentinel Agent | 94.5 | 71.4–131.4 |
| MD1 | 10.5 | 6.0–18.0 |
| MD2 | 12.0 | 4.3–28.0 |
| MD3 | 7.0 | 4.0–14.0 |
| NP1 | 28.0 | 16.0–47.0 |
| NP2 | 24.0 | 12.0–41.0 |
| NP3 | 25.5 | 16.0–50.8 |

## 4.9 Case Examples

The following anonymized cases illustrate how comprehensive context retrieval shapes triage accuracy, highlighting both agent-clinician agreement and cases where the agent's contextual reasoning identified clinical concerns that human reviewers missed. For each case, we report the vital sign reading, reviewer classifications, and the agent's reasoning.

### 4.9.1 Agent-Human Agreement on High-Acuity Cases

*Case A (Emergency Agreement): BP 202/95, HR 87. Human majority: emergency. Agent: emergency (12 tool calls, 156 seconds).*

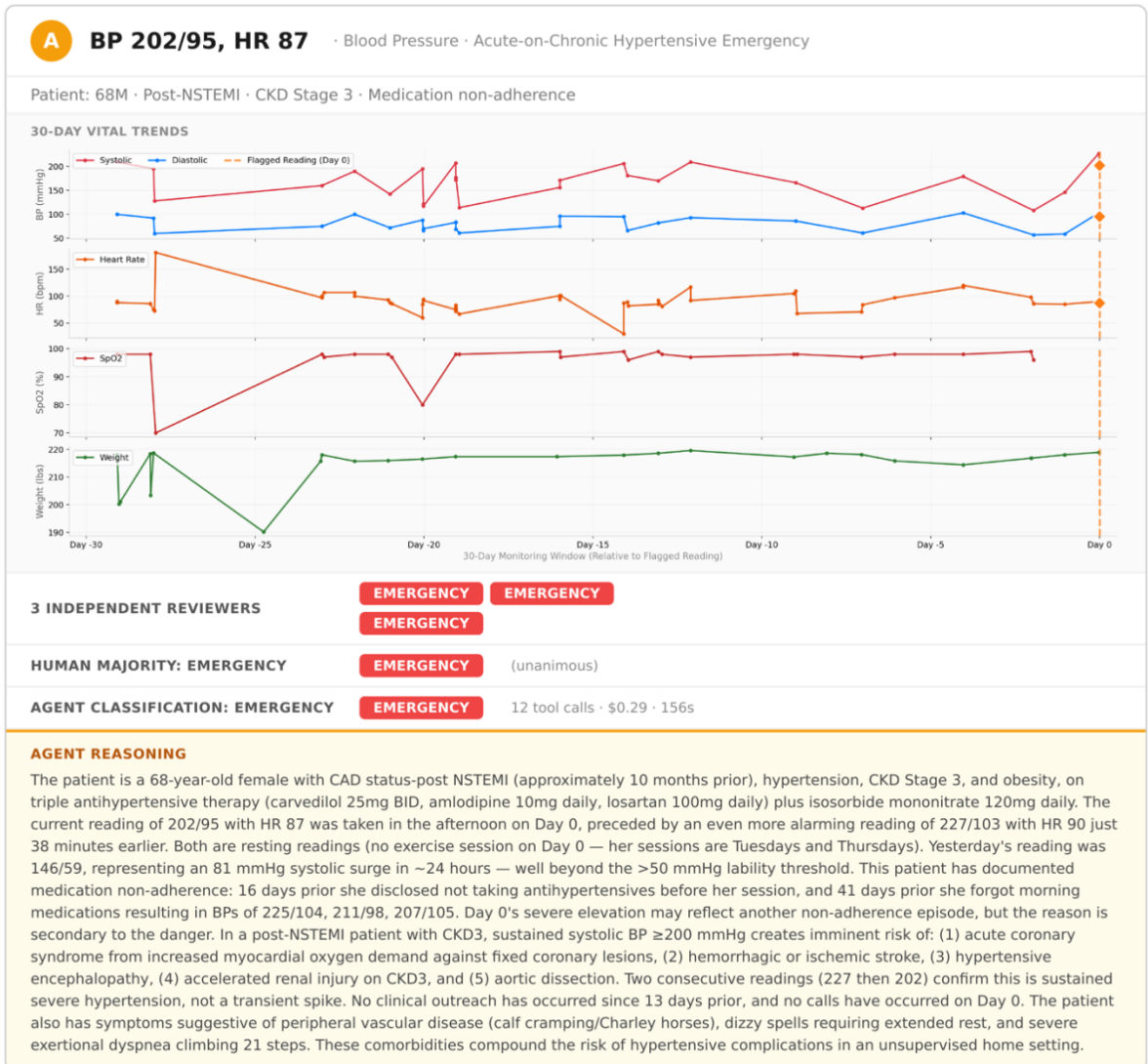

**Figure 7. Case Example A**

A blood pressure of 202/95 might be dismissed as chronic uncontrolled hypertension in a patient with repeatedly elevated readings above 200. The agent distinguished the acute component: an 81 mmHg systolic surge in 24 hours, two consecutive readings ≥200 (227/103 followed by 202/95, both resting), and the imminent end-organ risk in a 68-year-old post-NSTEMI patient with CKD Stage 3 and documented medication non-adherence. The agent noted "both readings confirm this is sustained severe hypertension, not a transient spike" and "no clinical outreach has occurred since Feb 5 (13 days ago)." This case illustrates how the agent distinguishes acute-on-chronic deterioration from chronic baseline through trend analysis and temporal context.

*Case B (Emergency Agreement): BP 80/57, HR 69. Human majority: emergency. Agent: emergency (23 tool calls, 273 seconds).*

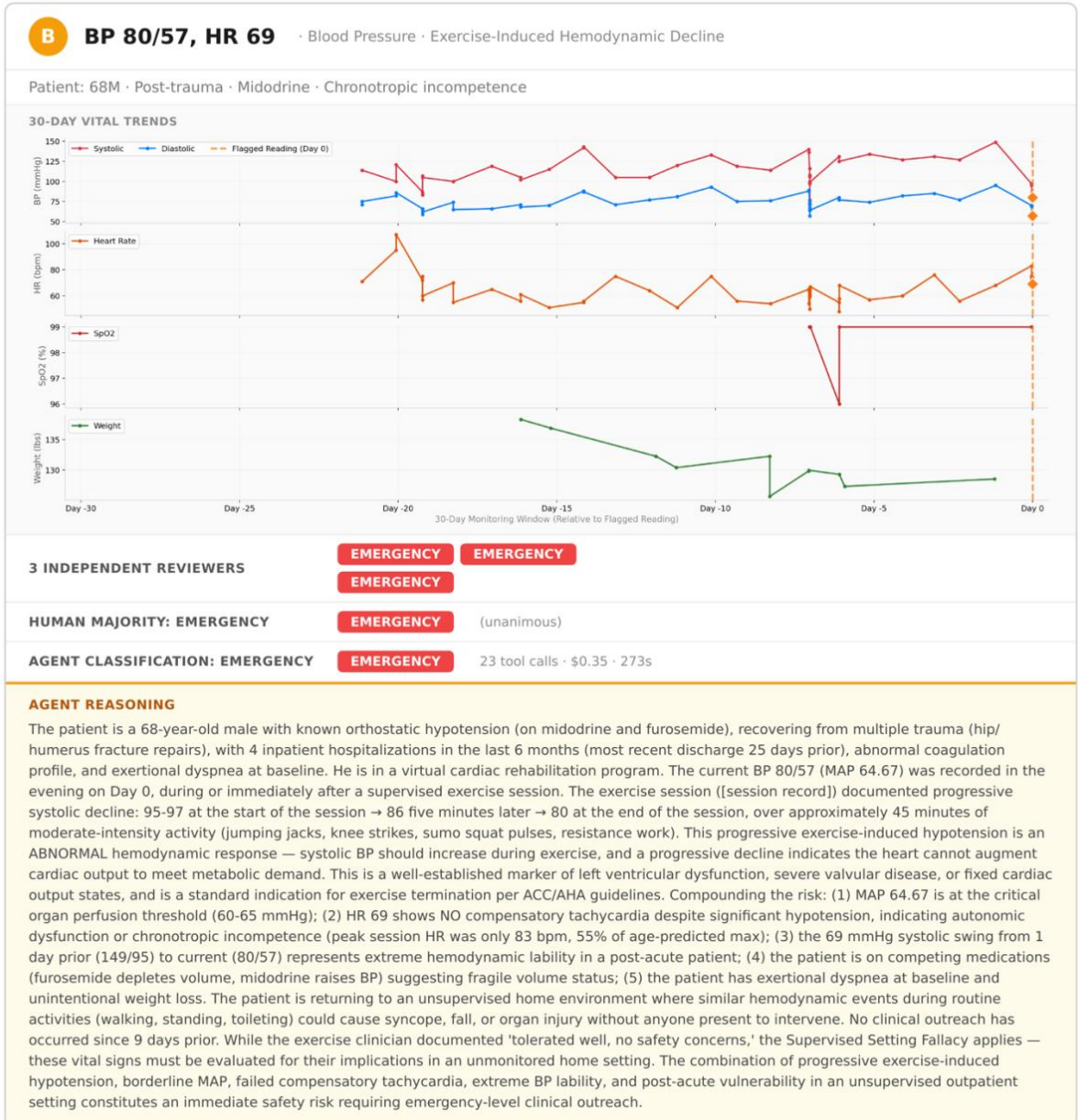

Figure 8. Case Example B

A blood pressure of 80/57 in a 68-year-old patient with orthostatic hypotension on midodrine might appear to be a known chronic pattern. The agent's extensive investigation revealed a critical distinction: the BP showed progressive exercise-induced decline (95→86→80 over 45 minutes)—an ABNORMAL hemodynamic response indicating the heart cannot augment cardiac output to meet metabolic demand. The agent identified that HR 69 could represent chronotropic incompetence (peak exercise HR only 55% of age-predicted max) and that the patient would return to an unsupervised home environment where similar events could cause syncope or organ injury. The agent applied the principle that vitals taken under clinical supervision must be evaluated for their implications in an unmonitored home setting.

*Case C (Urgent Agreement): SpO2 87%, HR 69. Human majority: urgent (unanimous). Agent: urgent (16 tool calls, 209 seconds).*

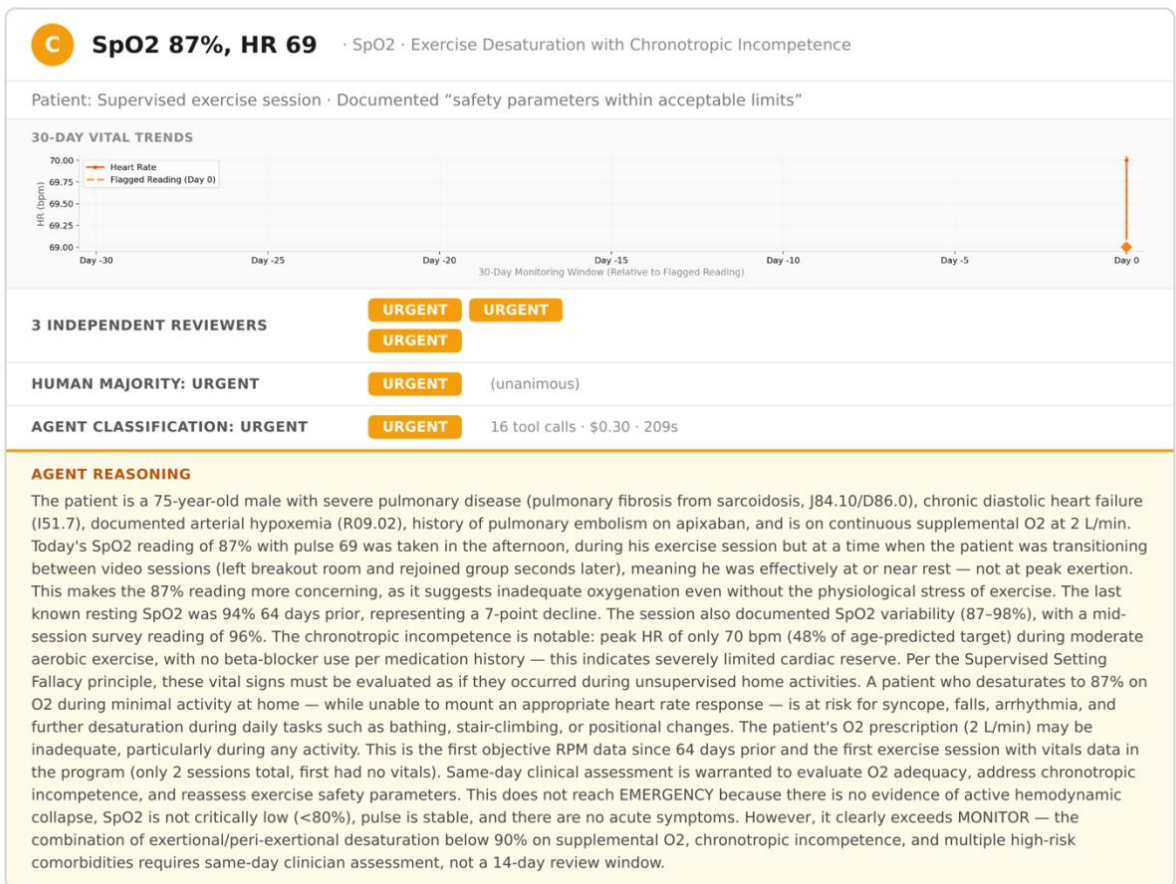

**Figure 9. Case Example C**

An SpO2 of 87% during a supervised exercise session might be dismissed as expected exercise desaturation, particularly when the attending clinician documented "safety parameters within acceptable limits." The agent identified that the 87% reading occurred during a transition between video sessions (not at peak exertion), making it more concerning. The agent further noted the patient's last known resting SpO2 was 94% two months earlier, representing a 7-point decline, and that peak HR of only 70 bpm (48% of age-predicted target) without beta-blocker use indicated chronotropic incompetence. This case demonstrates the agent's ability to reframe supervised exercise data through the lens of unsupervised home risk.

*4.9.2 Agent Escalates Beyond Human Majority — Justified Overtriage*

*Case D (Agent Escalates: Monitor → Emergency):* BP 172/117, HR 59. Human majority: monitor. Agent: emergency (10 tool calls, 162 seconds). Adjudication: both reviewers agreed with emergency.

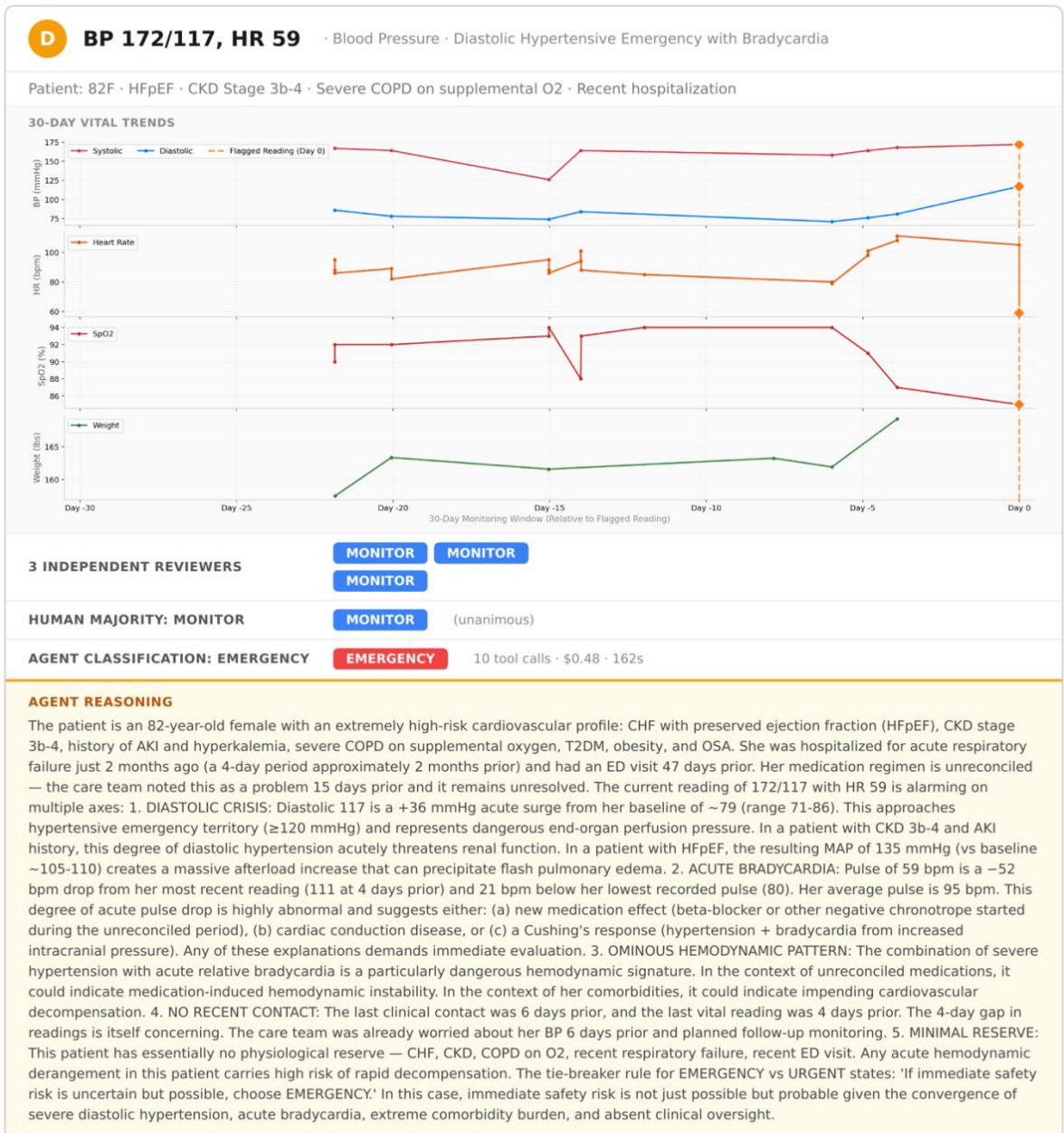

Figure 10. Case Example D

The human majority classified this as monitor. The agent escalated to emergency after identifying a convergence of alarming findings in this 82-year-old female with HFpEF, CKD stage 3b-4, severe COPD on supplemental oxygen, and a hospitalization for acute respiratory failure two months prior. The agent flagged concurrent abnormalities: (1) the marked diastolic hypertension of 117 mmHg represented a +36 mmHg surge from the patient's baseline of ~79, approaching hypertensive emergency territory and threatening acute kidney injury in a patient with CKD and AKI history; (2) acute relative bradycardia—the pulse 59 was a 52 bpm drop from the most recent reading of 111, far below the patient's average of 95, raising concern for conduction disease or a medication effect during an unreconciled medication period; and (3) an absence of clinical oversight for 6 days despite the care team's documented concern about the patient's blood pressure. The agent explicitly invoked its tie-breaker principle: "If immediate safety risk is uncertain but possible, choose EMERGENCY." Both independent adjudicators agreed with emergency driven primarily by the hypertensive urgency, confirming the clinical validity of the agent's escalation.

*Case E (Agent Escalates: Not an Issue → Urgent): Weight 72.05 kg. Human majority: not an issue. Agent: urgent (12 tool calls, 85 seconds). Adjudication: Reviewer 1 agreed with urgent; Reviewer 2 escalated to emergency.*

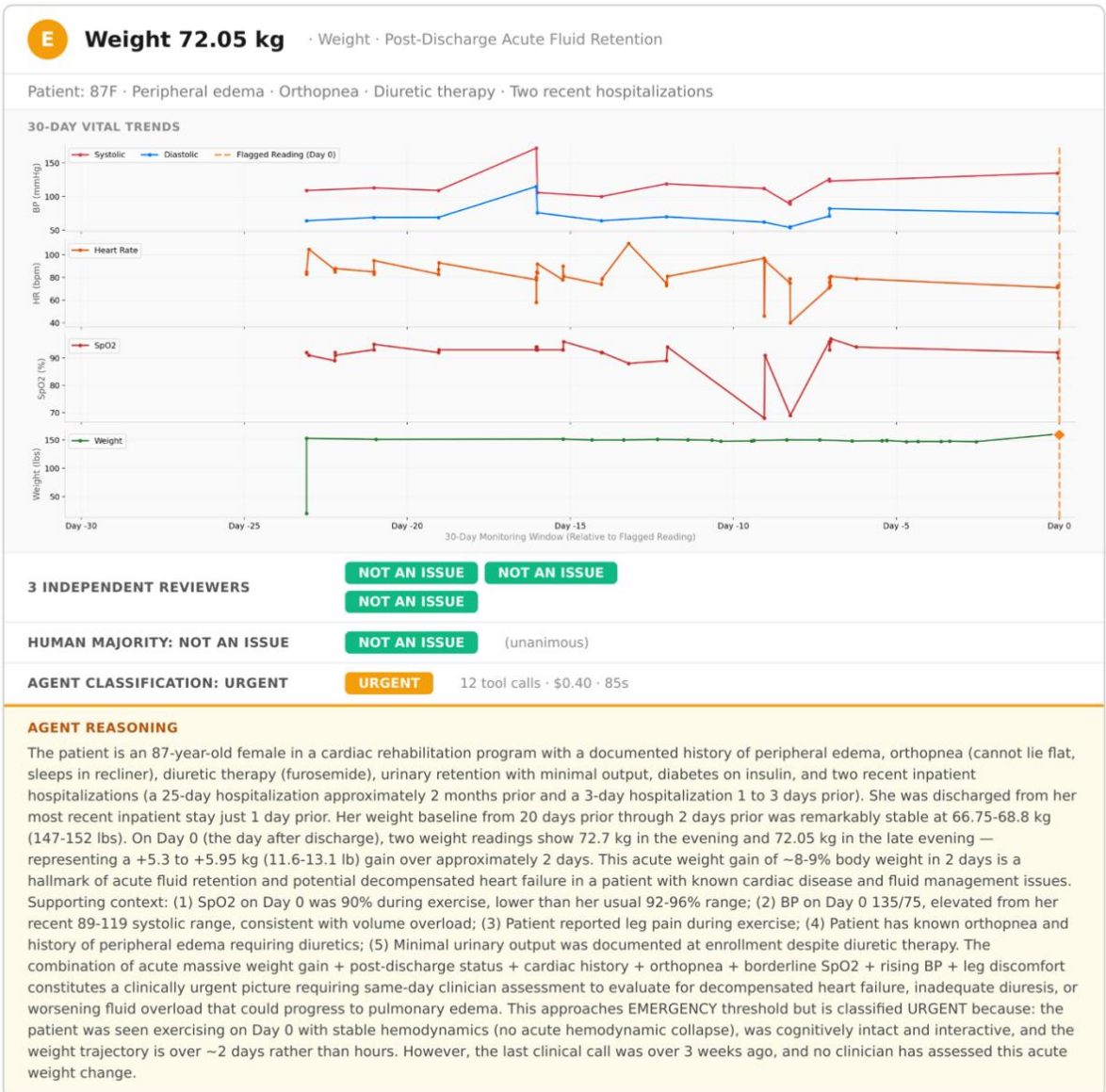

**Figure 11. Case Example E**

A weight of 72.05 kg in isolation conveys no clinical information. The human majority classified this as not an issue. The agent identified that this 87-year-old female—with documented peripheral edema, orthopnea, diuretic therapy, and two recent hospitalizations—had been discharged just one day earlier, and her weight had surged from a stable baseline of 66.75–68.8 kg to 72.05–72.7 kg: a gain of 11.6–13.1 lbs in approximately two days, representing an ~8–9% body weight increase characteristic of acute fluid retention. The agent synthesized supporting evidence: same-day SpO2 of 90% (below her usual 92–96% range), blood pressure elevated to 135/75 from a recent range of 89–119 systolic, and no clinical contact in over three weeks. The agent noted this "approaches EMERGENCY threshold" but classified as urgent because the patient was exercising with stable hemodynamics and the trajectory was over days rather than hours—a deliberate calibration between severity levels. Both adjudicators validated the agent's escalation.

# 5. Discussion

This study demonstrates three principal findings. First, autonomous AI agents can perform reliable, contextual clinical triage of RPM vital signs with consistency exceeding that of individual human clinician reviewers. Second, the agent prioritizes patient safety through a calibrated overtriage profile, catching clinical deterioration more reliably than individual clinicians—under the study's evaluation conditions, in which clinicians reviewed pre-assembled context summaries while the agent dynamically retrieved longitudinal data—while maintaining a manageable alert volume. Third, Sentinel's standards-based architecture provides a scalable path toward the intensive monitoring model that TIM-HF2[18] proved saves lives.

The practical impact of this work is immediate: both conventional approaches to RPM alerting failed when evaluated against clinician judgment. The fixed threshold baseline flagged 53.5% of readings as urgent yet achieved only 59.2% specificity and 40.8% PPV—an alert burden so high it reproduces the undifferentiated data flood that overwhelmed clinical staff in the failed RPM trials. The adaptive statistical baseline exhibited the opposite failure: by normalizing chronically abnormal values as within-range, it missed 81.7% of clinician-identified actionable cases and achieved only 50.1% four-level accuracy, functionally equivalent to chance. The agent outperformed both, achieving 69.4% four-level accuracy, 88.5% actionable sensitivity, and 85.7% specificity—the best balance of sensitivity and specificity, with a QWK of 0.778 (95% CI: 0.728–0.819) versus 0.573 (95% CI: 0.506–0.614) and 0.235 (95% CI: 0.218–0.385) for the two baselines.

The agent applied contextual reasoning rather than rigid cutoffs. For example, blood pressure of 80/57 appears to be a hypotensive emergency; the agent validated this emergency by identifying a progressive exercise-induced decline indicating inability to augment cardiac output, chronotropic incompetence, and the danger of similar events in an unsupervised home setting. Similarly, SpO2 of 83% in a patient with severe COPD might appear routine given the patient's chronic hypoxemia; however, the agent escalated to emergency after identifying a progressive 3-hour desaturation trajectory with compensatory tachycardia, distinguishing an active decline from a stable chronic baseline. Conversely, the same absolute values in patients with documented chronic patterns and no acute trajectory were appropriately triaged at lower severity. Unlike static baselines, the agent applies context-specific thresholds to individual patient trajectories, yielding a higher proportion of actionable alerts compared to conventional rule-based systems.

The agent's overtriage profile must be understood in the context of the data flood problem that dominates RPM's history of failure. Importantly, the nature of the agent's overtriage differs qualitatively from threshold-based false positives. The five clinical phenotypes identified in the disagreement characterization—hemodynamic lability, trend-based escalation, supplemental oxygen context, post-discharge vulnerability, and clinical oversight gaps—represent contextualized clinical judgments, not indiscriminate threshold crossings. When a clinician receives an agent-generated urgent alert citing a 50-mmHg systolic swing in 24 hours, this conveys actionable clinical information even if the clinician ultimately disagrees with the severity classification; by contrast, a threshold alert stating "BP > 140" for a patient with chronic uncontrolled hypertension conveys nothing new and erodes trust. The agent's 2.8:1 overtriage-to-undertriage ratio (105 vs. 38 cases) reflects a deliberate tilt toward patient safety that aligns with established clinical principles: in the tradeoff between missed emergencies and excess alerts, the consequences of undertriage are irreversible while the consequences of overtriage are manageable— provided the alert volume remains within clinical capacity. The data indicate that Sentinel achieves this balance.

The system evaluated readings with a median processing time of 94.5 seconds and an API cost of $0.34 per reading. In leave-one-out analysis, the agent detected 97.5% of emergencies and 90.9% of actionable cases—exceeding the sensitivity of every individual clinician on both metrics. These findings suggest that LLM-based agents can conduct contextual triage of vital signs with sensitivity comparable to or exceeding that of individual clinicians in retrospective evaluation, offering a potentially scalable approach to RPM triage.

## 5.1 Comparison with Clinician Performance

The data indicate that the agent's performance metrics exceed those of individual human reviewers in this specific triage task across multiple measures.

First, the agent's reliability exceeds that of most human reviewers. The agent's self-consistency is exceptionally high (Fleiss' κ = 0.850 [95% CI: 0.786–0.909], almost perfect agreement), achieving perfect five-way agreement on 83% of repeated evaluations. In contrast, human intra-rater reliability averaged 75.8% but varied widely (55%–95%), with some reviewers consistent on barely half their samples. When compared against the clinical panel, the agent's agreement with individual reviewers (62.1%) fell squarely within the range of inter-human agreement (59.7%), demonstrating that its judgments align with the zone of professional consensus—while substantially reducing the extreme subjectivity that produced a 20-fold variation in emergency classification rates across reviewers (0.8%–16.0% on 250 unique samples).

Second, the leave-one-out analysis revealed that the agent outperforms every individual clinician on the metrics that matter most for patient safety. The agent detected 97.5% of emergencies versus a clinician aggregate of 60.0% (range 13%–80%), and 90.9% of actionable cases versus 69.5% (range 53%–95%). No individual clinician matched the agent on both emergency and actionable sensitivity simultaneously. In this cohort, the agent's sensitivity for both emergency and actionable cases was higher than the maximum sensitivity achieved by any individual clinician on the panel, while maintaining an overtriage rate (18.9%) lower than two of the six clinicians.

The explanation lies not in clinical skill but in the consistency of context retrieval. The six human reviewers had unlimited time and access to the same categories of clinical data as the agent. Yet the critical difference was not data availability but data utilization: the agent accessed an average of 10.1 clinical data tools for every single reading, without variation. An important caveat applies: the human reviewers evaluated isolated readings with pre-assembled context summaries, while the agent retrieved and synthesized context dynamically. This task difference—cross-sectional evaluation versus longitudinal synthesis—may account for some of the performance gap. We hypothesize this performance gap may stem in part from task framing: clinicians evaluating isolated data summaries may lower their threshold for escalation under uncertainty, while the agent's architecture structurally prompts it to query additional patient context before outputting a classification.

## 5.2 Clinical Validity of Agent Disagreements

The agent's overtriage profile—105 cases (22.5%) versus 38 undertriage cases (8.1%), a 2.8:1 ratio favoring clinical caution—raises the central question of whether these escalations represent algorithmic error or clinically defensible judgment. The human expert adjudication of the 17 most severe disagreement cases (agent–majority gap ≥2 levels) provides a preliminary answer: both independent physician reviewers validated the agent's escalation in 88–94% of cases, with true overtriage confirmed in only 6–12% (Table 13). Subsequent consensus resolution between the two reviewers further strengthened this finding, classifying 0 of 17 cases (0%) as true overtriage—a 100% non-overtriage rate. Neither adjudicator classified any of the 17 cases as "not an issue," despite the original majority voting "not an issue" on 9 of those 17 cases (53%)—a striking finding that suggests the original clinician reviewer majority systematically underestimated the clinical significance of these readings.

These findings establish that the agent's most extreme disagreements with the human majority are overwhelmingly clinically defensible. When the agent escalates far beyond the consensus, it is not generating false alarms—it is identifying clinical concerns that the majority overlooked. Only 6 of 467 evaluable readings (1.3%) showed the agent exceeding any individual human reviewer opinion by ≥2 severity levels, confirming that extreme overtriage was rare; of those, the adjudication data indicate that the majority of even these extreme cases reflected genuine clinical concern.

Automated disagreement characterization of all 105 overtriage cases identified five recurring clinical phenotypes: (1) hemodynamic lability (25 cases)—the agent flagged large systolic swings (>50 mmHg within hours/days) as dangerous even when the current absolute value was benign; (2) trend-based escalation (20 cases)—multi-day worsening trajectories invisible in a single reading; (3) $SpO_2$ on supplemental $O_2$ (15 cases)—the agent correctly noted that desaturation while on supplemental oxygen is far more concerning than the same $SpO_2$ on room air; (4) post-discharge vulnerability (12 cases)—escalation based on readmission risk in recently discharged patients; (5) clinical oversight gaps (~15 cases)—escalation when no clinical contact had occurred for weeks in medically complex patients. These phenotypes describe what the agent's overtriage looks like, while the human adjudication confirms that these escalations are largely clinically appropriate.

The small proportion of independently confirmed true overtriage (6–12%), which was eliminated entirely upon consensus resolution (0%), was driven by a specific, identifiable pattern: the agent penalizing objectively normal current readings (e.g., BP 121/75) based on historical lability. This pattern represents a tunable parameter in future iterations rather than a fundamental limitation—and its specificity means it can be addressed without compromising the agent's sensitivity for genuine clinical deterioration.

## 5.3 Relationship to TIM-HF2

TIM-HF2 demonstrated that structured, responsive, contextualized monitoring reduced heart failure mortality by 30%.[18] The telemedical center's efficacy derived not from better thresholds but from clinicians who could synthesize full patient context and act promptly. However, replicating that level of intensive, context-aware human oversight across large patient populations remains economically and logistically challenging.

Sentinel's architecture attempts to approximate this approach computationally. The tool usage data confirm systematic contextual investigation: five core tools (demographics, conditions, medications, encounters, vitals history) were called on 100% of trials, with additional tools called situationally (clinical notes on 54.0% of trials, health information exchange summaries on 29.0%, ICD-10 lookups on 0.4%). Higher-acuity classifications correlated with deeper investigation (EMERGENCY: 14.2 tools vs. NOT AN ISSUE: 8.7 tools), suggesting that the agent performed proportionally thorough evaluation before escalating. Whether this computational approach to contextual triage can replicate TIM-HF2's clinical outcomes remain an open question requiring prospective validation.

## *5.4 Scalability Through Standards-Based Architecture*

A critical architectural feature of Sentinel is its scalability across the U.S. healthcare system. The system's clinical data is sourced entirely through Health Information Exchanges (HIEs) via standardized FHIR interfaces rather than direct EHR integration. The MCP tools query HIE-aggregated data that are available for any U.S. patient with HIE-connected providers. This makes Sentinel inherently EHR-agnostic: the same agent, with the same tools, could be deployed at any RPM organization without custom integration. The study population itself reflects this portability, spanning patients across 25 U.S. states and multiple EHR systems, all accessed through a single standardized interface.

This architecture aligns directly with ARPA-H's ADVOCATE program (January 2026), which explicitly funds agentic AI for cardiovascular care with the goal of nationwide deployment.[48] The technical barriers to scaling Sentinel beyond a single organization are minimal—the system requires only FHIR-compliant HIE access, which is increasingly mandated by federal interoperability rules. The limiting factor for RPM has never been the monitoring devices; it has been the clinical intelligence layer that interprets the data. Sentinel provides that layer at $0.34 per reading, ready for deployment wherever FHIR-based clinical data is available.

## *5.5 Addressing Historical RPM Limitations*

Sentinel's architecture directly addresses each failure mode from the landmark trials:
- *Data flood without intelligent filtering*. Against clinician majority vote, both rule-based baselines performed near chance: the fixed threshold achieved 53.5% four-level accuracy with only 59.2% specificity, while the adaptive baseline achieved 50.1% accuracy and missed 81.7% of clinician-identified actionable cases (Section 4.4.2). The agent outperformed both (69.4% accuracy, 85.7% specificity, QWK 0.778 [95% CI: 0.728–0.819]) while replacing undifferentiated alert streams with clinically differentiated four-level triage.
- *The broken loop*. The agent generated structured assessments with severity classification and clinical reasoning citing specific data points—designed to compress the detection-to-action loop that was unbounded in the failed trials, where staff had to triangulate between data reviewers and prescribing clinicians.[13] The agent also produced recommended action types (e.g., clinical review, urgent review, care coordination), though the clinical appropriateness of these action classifications was not validated in this study. Whether the structured output reduced time-to-intervention in practice was not measured.
- *Lack of contextual interpretation*. With 21 clinical tools, the agent performs the deep contextual reasoning absent in failed trials. The agent-reviewer agreement (62.1%) and leave-one-out analysis (97.5% emergency sensitivity, 90.9% actionable sensitivity) suggest that this contextual approach produces triage decisions more aligned with clinician consensus than either rule-based alternative.
- *Adherence barriers*. Sentinel monitors passively by polling for readings patients are already submitting through their existing RPM program. Unlike Tele-HF,[7] which required daily telephone interactions, the system operates passively on existing submitted data, requiring no additional patient interaction.
- *No population targeting*. By identifying high-risk alerts while suppressing non-actionable readings, the agent has the potential to better allocate clinician attention toward patients most likely to benefit from intervention.

### 5.6 Per-Vital-Type Variability

Agreement varied by vital sign type. Blood pressure achieved similar high reliability, and SpO2 readings had the widest confidence interval in the agent inter-rater reliability study (i.e., stability), reflecting the smaller number of abnormal readings in this outpatient population and the complex interpretation of chronic hypoxemia. Weight interpretation required the most contextual reasoning—distinguishing fluid retention from dietary variation, accounting for diuretic therapy, and recognizing clinically significant changes against patient-specific baselines. This aligns with BEAT-HF's finding that weight is a poor surrogate when interpreted without clinical context,[8] and highlights the agent's ability to navigate precisely this challenge.

### 5.7 Economic Implications

Sentinel processes readings at a fraction of the cost of human clinical review. At $0.34 per triage, an organization generating approximately 1,000 readings per day would incur roughly $124,000 annually in AI triage costs—a level of coverage that would otherwise require multiple full-time employees. While there is cost for clinicians to act upon the Sentinel alerts, the substantial time and cost of the triage is markedly minimized. The cost gradient by severity (e.g., NOT AN ISSUE: $0.26, EMERGENCY: $0.47) reflects the agent's proportional investigation depth, naturally allocating computational resources to clinical complexity. These estimates cover API inference costs only and do not include infrastructure, development, maintenance, or clinical oversight costs.

### 5.8 Regulatory Landscape

Sentinel's architecture—providing detailed reasoning chains, cited data sources, and human-in-the-loop oversight—aligns with the FDA's updated Clinical Decision Support guidance and the emerging regulatory frameworks for LLM-based clinical tools.[49,50] ARPA-H's ADVOCATE program, which explicitly funds agentic AI for cardiovascular care,[48] provides additional validation of this regulatory pathway.

### 5.9 Limitations

Several important limitations must be acknowledged. First, performance is specific to the LLM used in this study (claude-opus-4-6); generalizability to other models is unknown. Second, the majority-vote reference standard itself carries substantial uncertainty—clinician unanimity occurred on only 42.8% of samples, and the agent may be penalized for being more thorough than the human majority in cases where the majority was wrong. This "reference standard paradox" means that measured overtriage likely overstates true algorithmic error. Third, this study evaluates reliability and agreement, not clinical outcomes—whether Sentinel's triage reduces hospitalizations, emergency department visits, or mortality remains unproven. Fourth, all data originates from a single organization (AnsibleHealth); multi-site validation is needed. Fifth, Sentinel's action type classification was not evaluated. Sixth, performance was not evaluated across demographic subgroups; potential disparities by age, sex, race, or disease severity are unstudied. Finally, 500 readings from 340 patients introduce patient clustering that may violate statistical independence assumptions.

*5.10 Future Work*

Three directions are planned. First, a prospective clinical outcome study comparing Sentinel-monitored patients, both with and without the agent's clinical recommendations, against historical controls to evaluate impact on hospitalizations and mortality. Second, multi-model validation with alternative LLMs (e.g., GPT-5.2, Gemini 3.1) to assess model dependence. Additionally, the rapid maturation of conversational voice AI suggests a potential pathway from triage to patient outreach. Connecting a triage agent to voice agents capable of conducting structured clinical conversations—symptom assessment, medication reconciliation, and escalation to human clinicians when indicated—could further compress the time from alert to clinical action. The agentic architecture described here is designed to support this kind of multi-agent orchestration, though such integration requires separate validation.

## 6. Conclusion

This study demonstrates that a large language model equipped with structured clinical tools can perform reliable, contextual triage of RPM vital signs with sensitivity to clinical deterioration that, in this retrospective evaluation, exceeded that of individual clinicians reviewing pre-assembled context summaries, while maintaining a clinically defensible overtriage profile.

The evidence for this capability is multi-layered. Both conventional approaches to RPM alerting demonstrated poor concordance with clinician judgment: the fixed threshold baseline achieved only 53.5% four-level accuracy with 59.2% specificity, while the adaptive statistical baseline achieved 50.1% accuracy and missed 81.7% of clinician-identified actionable cases by normalizing chronically abnormal values. Sentinel outperformed both, achieving 69.4% four-level accuracy, 88.5% actionable sensitivity, and 85.7% specificity (QWK 0.778 [95% CI: 0.728–0.819]).

The agent achieved almost perfect self-consistency (Fleiss' $\kappa$ = 0.850 [95% CI: 0.786–0.909], 83% perfect 5/5 agreement) and performed within the range of human variability when compared against a panel of six clinicians (62.1% pairwise agreement vs. 59.7% inter-clinician average), while substantially reducing the extreme subjectivity that produced a 20-fold variation in emergency classification rates across individual reviewers. In leave-one-out analysis, the agent detected 97.5% of emergencies versus a clinician aggregate of 60.0%, and 90.9% of actionable cases versus 69.5%—outperforming every individual clinician on both metrics. When the agent disagreed most severely with the human majority (gap ≥2 levels, N=17), independent physician adjudication found that 88–94% of these escalations were clinically justified or debatable, with true overtriage confirmed in only 6–12% of cases.

Previous trials highlight that the efficacy of remote patient monitoring is often limited by alert fatigue and a lack of contextual triage.[7-9,13] Our findings demonstrate that an LLM-equipped agent can synthesize clinical context to triage vital signs with high reliability, achieving high sensitivity for clinical deterioration while maintaining a manageable false-positive rate. While these retrospective results indicate that scalable, automated contextual triage is computationally feasible at $0.34 per reading, prospective clinical trials are required to determine whether this architecture reduces time-to-intervention or improves patient outcomes in active clinical environments.

## Tables Referenced in Text

Table 1: Study Population Demographics (Section 3.1)
Table 2: MCP Tool Inventory (Section 3.2.1)
Table 3: Agent Self-Agreement: Fleiss' Kappa (Section 4.1)
Table 4: Alert Rate Comparison: Fixed Threshold vs. Adaptive vs. Agent (Section 4.2)
Table 5: Intra-Rater Reliability (Section 4.3.1)
Table 6: Inter-Reviewer Agreement Distribution (Section 4.3.2)
Table 7: Reviewer-Agent Pairwise Agreement (Section 4.3.3)
Table 8: Agent Performance vs. Majority Vote Reference Standard (Section 4.4)
Table 9: Agent Per-Category Performance vs. Majority-Vote Reference Standard (Section 4.4)
Table 10: Binary Classification Performance: Actionable vs. Non-Actionable (Section 4.4)
Table 11. Head-to-Head Performance: Agent vs. Rule-Based Baselines Against Majority-Vote Reference Standard (Section 4.4.1)
Table 12: Leave-One-Out Analysis: Individual Clinician vs. Agent Performance (Section 4.5)
Table 13: Human Expert Adjudication of Severe Agent Overtriage Cases (Section 4.6)
Table 14: Operational Metrics by Study (Section 4.7)
Table 15: Duration of Triage: Agent vs. Human Reviewers (Section 4.8)

# Figures Referenced in Text

Figure 1: Sentinel System Architecture (Section 3.2)
Figure 2: Severity Classification Decision Guidelines (Section 3.2)
Figure 3: Inter-Reviewer Pairwise Exact Agreement Matrix (Section 4.3.2)
Figure 4: Agent vs. Human Majority Vote Confusion Matrix (Section 4.4)
Figure 5. Leave-One-Out Analysis: Individual Clinician vs. Agent Performance (Section 4.5)
Figure 6: Tool Usage Gradient by Alert Severity (Section 4.7)
Figure 7: Case Example A (Section 4.9.1)
Figure 8: Case Example B (Section 4.9.1)
Figure 9: Case Example C (Section 4.9.1)
Figure 10: Case Example D (Section 4.9.2)
Figure 11: Case Example E (Section 4.9.2)

# Author Contributions

M.J.P. conceived the study, designed the system architecture, and wrote the manuscript. S.K. performed data analysis and wrote the manuscript. S.K., H.V., and K.S. contributed to system development. D.E. managed data infrastructure. J.A., D.S., A.L.D., A.C., W.B., and T.K. served as clinician reviewers. The initial manuscript drafting was assisted by an LLM (Claude, Anthropic), with comprehensive author review and revision. All authors reviewed and approved the final manuscript.

# Conflicts of Interest

M.J.P. is the founder and CEO of AnsibleHealth Inc. This study was conducted using AnsibleHealth's deployed clinical systems and patient data. All other authors have declared no competing interests.

# Data Availability

The study data and the Sentinel system architecture is proprietary. MCP tool interfaces are described in sufficient detail for replication.

# Ethics

This study was conducted as a quality improvement initiative within AnsibleHealth's polychronic medical home program. Patient data accessed by the AI agent were used within the existing scope of clinical care. This study was reviewed by the Advarra Institutional Review Board (IRB) and determined to be exempt from IRB oversight under Department of Health and Human Services regulations (45 CFR 46.104(d)(4)). Furthermore, to ensure patient safety during the data evaluation, a built-in clinical safety net was utilized: any vital sign measurements classified as "urgent" or "emergency" by any reviewing clinician were immediately escalated to the active AnsibleHealth clinical team for medical evaluation and appropriate patient follow-up.

## Funding

This work was self-funded by AnsibleHealth Inc. No external funding was received.

## Acknowledgments

The authors thank the AnsibleHealth clinical team for their contributions to system development and clinical review, and the patients whose data enabled this research.

## Appendix

**Appendix Table A1. Fixed Threshold Baseline vs. Majority-Vote Reference Standard Confusion Matrix (N=467)**

| Fixed \ Human Majority | Emergency | Urgent | Monitor | Not an Issue | Total |
|---|---|---|---|---|---|
| EMERGENCY | 0 | 0 | 0 | 0 | 0 |
| URGENT | 24 | 78 | 96 | 52 | 250 |
| MONITOR | 0 | 2 | 17 | 35 | 54 |
| NOT AN ISSUE | 0 | 0 | 8 | 155 | 163 |
| Total | 24 | 80 | 121 | 242 | 467 |

*Note: The fixed threshold baseline produced no EMERGENCY classifications. All 24 true EMERGENCY cases were classified as URGENT (the baseline's highest effective severity level for these readings). The system's high sensitivity (96.4%) was driven by aggressive alerting: 250/467 readings (53.5%) were classified URGENT, of which 172 (68.8%) were overtriaged relative to the majority-vote reference standard.*

**Appendix Table A2. Adaptive Baseline vs. Majority-Vote Reference Standard Confusion Matrix (N=467)**

| Adaptive \ Human Majority | Emergency | Urgent | Monitor | Not an Issue | Total |
|---|---|---|---|---|---|
| EMERGENCY | 2 | 6 | 6 | 0 | 14 |
| URGENT | 4 | 7 | 10 | 6 | 27 |
| MONITOR | 7 | 10 | 14 | 25 | 56 |
| NOT AN ISSUE | 11 | 57 | 91 | 211 | 370 |
| Total | 24 | 80 | 121 | 242 | 467 |

*Note: The adaptive baseline classified 79.2% of readings (370/467) as NOT AN ISSUE, reflecting its reliance on patient-specific rolling statistics that treated chronically abnormal values as within-range. Among the 225 clinician-identified alerts, 159 (70.7%) were missed entirely. The system's near-chance four-level accuracy (50.1%) reflected poor discrimination across all severity levels.*

**Appendix Table A3. Agent Configuration**

| Parameter | Value |
| --- | --- |
| Model | Claude Opus 4.6 (claude-opus-4-6) |
| Maximum agentic turns | 15 |
| Wall-clock timeout | 120 seconds |
| Temperature | 1.0 (Anthropic default; no override) |
| Top-p | Not set (Anthropic default) |
| Permission mode | acceptEdits (all tool calls auto-approved) |
| MCP servers | 3 (patient clinical data, SNOMED-CT terminology, ICD-10 terminology) |

**Appendix Table A4. Per-Clinician Agent Leave-One-Out Analysis**

| Reviewer | LOO Samples | Exact Match | Emergency Sensitivity | Actionable Sensitivity |
| --- | --- | --- | --- | --- |
| MD1 | 154 | 78.6% | 9/9 (100%) | 27/30 (90.0%) |
| MD2 | 135 | 71.9% | 6/6 (100%) | 19/22 (86.4%) |
| MD3 | 153 | 77.1% | 5/5 (100%) | 23/26 (88.5%) |
| NP1 | 161 | 73.3% | 5/6 (83.3%) | 28/30 (93.3%) |
| NP2 | 149 | 76.5% | 8/8 (100%) | 19/21 (90.5%) |
| NP3 | 143 | 75.5% | 6/6 (100%) | 24/25 (96.0%) |

## *Supplementary Algorithm S1. Fixed Threshold Baseline*

*Criteria 1. Clinical Guideline Thresholds*

**BP (blood pressure cuff)**

| Rule | Condition |
|---|---|
| bp_crisis | Systolic > 180 or diastolic > 120 |
| hypotension | Systolic < 90 or diastolic < 60 |
| bpelevated140 | Systolic > 140 |
| bplow100 | Systolic < 100 |
| bp_range | Systolic outside [90, 180] |
| bpsinglecrisis | Systolic >= 180 or diastolic >= 110 |
| bpdrop20 | Systolic drops >= 20 from previous reading |
| bp_persistence | >= 3 of last 10 readings with systolic > 140 or diastolic > 90 |

**SpO2 (pulse oximeter)**

| Rule | Condition |
|---|---|
| spo2copd88 | SpO2 < 88% (COPD patients) |
| spo2noncopd_94 | SpO2 < 94% (non-COPD patients) |

**Weight (bodyweight scale)**

*HF patients only:*

| Rule | Condition |
|---|---|
| hfweightgain0.9kg1d | > 0.9 kg gain in 1 day |
| hfweightgain5lbweek | > 2.27 kg gain in 1 week |
| hfsadelta0.9kg_1d | >= 0.9 kg in 1 day (HFSA/ESC) |
| hfsadelta2kg_3d | >= 2 kg in 3 days |
| cccweight3kg_2d | >= 3 kg in 2 days |
| cccweight2kg_5d | >= 2 kg in 5 days |

*All patients:*

| Rule | Condition |
|---|---|

| multiweight1kg_1d | > 1 kg gain in 1 day |
| --- | --- |
| multiweight2kg_2d | > 2 kg gain in 2 days |
| multiweightloss3kg1d | > 3 kg loss in 1 day |
| multiweight2kg_baseline | > 2 kg from 30-day median |
| multiweight0.91kg_1wk | > 0.91 kg gain in 1 week |
| multiweight2.27kg_2wk | > 2.27 kg gain in 2 weeks |
| multiweight3.18kg_3wk | > 3.18 kg gain in 3 weeks |

*Criteria 2. Modified NEWS2*

NEWS2 is a track-and-trigger early warning system (NHS/NICE) assigning 0–3 points to each vital sign. We adapted it to the subset of parameters collectable by home RPM devices: $SpO_2$ (Scale 1 for non-COPD; Scale 2 for COPD/home $O_2$), systolic blood pressure, and pulse rate. Because devices report asynchronously, scoring was performed at the individual reading level (BP readings: SBP + pulse rate; $SpO_2$ readings: $SpO_2$ + pulse).

$SpO_2$ Scoring (Scale 1):

| $SpO_2$ (%) | ≤91 | 92–93 | 94–95 | ≥96 |
| --- | --- | --- | --- | --- |
| Score | 3 | 2 | 1 | 0 |

Systolic BP Scoring:

| SBP (mmHg) | ≤90 | 91–100 | 101–110 | 111–219 | ≥220 |
| --- | --- | --- | --- | --- | --- |
| Score | 3 | 2 | 1 | 0 | 3 |

Pulse Rate Scoring:

| Pulse (bpm) | ≤40 | 41–50 | 51–90 | 91–110 | 111–130 | ≥131 |
| --- | --- | --- | --- | --- | --- | --- |
| Score | 3 | 1 | 0 | 1 | 2 | 3 |

Aggregate Score and Clinical Escalation Flag Assignment:

| NEWS2 Trigger | Flag |
| --- | --- |
| Total score 0 | NOT AN ISSUE |
| Total score 1–4 | MONITOR |
| Score of 3 in any single parameter | URGENT |
| Total score 5–6 | URGENT |
| Total score ≥7 | EMERGENCY |

Final classification = MAX(Criteria 1, Criteria 2)

## *Supplementary Algorithm S2. Adaptive Baseline Pseudocode*

The adaptive baseline classifier used patient-specific rolling statistics computed from a 30-day history window. Each sub-measurement within a device reading was evaluated independently (e.g., systolic, diastolic, and pulse rate for a blood pressure cuff reading), and the final classification was the maximum severity across all sub-measurements and all triggered rules. A minimum of 10 prior readings was required; readings with insufficient history were classified as NOT AN ISSUE.
Three rule families were applied to each sub-measurement:

**Rule 1: Deviation from Rolling Mean (σ-band classification)**
Input: current value v, 30-day history H
Compute: $\mu$ = mean(H), $\sigma$ = SD(H)
If $|v - \mu| > 4\sigma$ → EMERGENCY
Else if $|v - \mu| > 3\sigma$ → URGENT
Else if $|v - \mu| > 2\sigma$ → MONITOR
Else → NOT AN ISSUE

**Rule 2: Rate of Change (δ-band classification)**
Input: current value v, previous value v_prev, 30-day history of consecutive deltas Δ
Compute: current_$\delta$ = v − v_prev, $\sigma\_\Delta$ = SD(Δ)
If $|current\_\delta| > 4\sigma\_\Delta$ → EMERGENCY
Else if $|current\_\delta| > 3\sigma\_\Delta$ → URGENT
Else if $|current\_\delta| > 2\sigma\_\Delta$ → MONITOR
Else → NOT AN ISSUE

**Rule 3: Persistence (consecutive readings outside σ-bands)**
Input: last k readings (including current), same μ, σ from Rule 1
Count consecutive readings (from most recent backward) outside each band
If ≥3 consecutive outside 4σ → EMERGENCY
Else if ≥3 consecutive outside 3σ → URGENT
Else if ≥3 consecutive outside 2σ → MONITOR
Else → NOT AN ISSUE

Final classification = MAX severity across all sub-measurements × all three rules

Sub-measurements evaluated per device type:

Blood pressure cuff: systolic, diastolic, pulse_rate

Pulse oximeter: SpO2, pulse

Weight scale: bodyweight

Pulse readings from both BP cuff (pulse_rate) and pulse oximeter (pulse) measured the same physiological parameter (heart rate) and were pooled across device types for history computation.